\documentclass[journal, twoside]{IEEEtran}
\ifCLASSINFOpdf
  \usepackage[pdftex]{graphicx}
  \graphicspath{{figures/}}
\else
\fi
\usepackage{amsmath}

\usepackage[utf8]{inputenc}
\usepackage{subcaption} 
\usepackage{hyperref}
\usepackage{multirow} 

\usepackage{mathtools} 
\usepackage{amssymb}
\usepackage{amsthm}
\usepackage{mathtools}
\usepackage{stmaryrd}
\usepackage[binary-units=true]{siunitx}
\usepackage{chngcntr}
\usepackage{algorithm}
\usepackage{algpseudocode}
\usepackage{tikz}
\usepackage{pgfplots}
\usepackage{multirow}
\usepackage{booktabs}
\usepackage{calculator}
\usetikzlibrary{intersections}
\pgfplotsset{compat=newest}
\usepgfplotslibrary{fillbetween}

\def \O {\mathcal{O}}
\def \argmax {\text{argmax}}

\begin{document}
\bstctlcite{IEEEexample:BSTcontrol}
%
\title{Scalable Robust Graph and Feature Extraction for Arbitrary Vessel Networks in Large Volumetric Datasets}
%
%
%

\author{
  Dominik~Drees,
  Aaron Scherzinger,
  René Hägerling,
  Friedemann Kiefer,
  Xiaoyi Jiang,~\IEEEmembership{Senior Member,~IEEE}
}

\maketitle


\let\svthefootnote\thefootnote
\let\thefootnote\relax\footnotetext{
  D.\ Drees and X.\ Jiang are with the Faculty of Mathematics and Computer Science, University of Münster, Germany.\\
  A.\ Scherzinger is with Ubisoft, Mainz, Germany.\\
  R.\ Hägerling is with Charité -- Universitätsmedizin Berlin, Germany.\\
  F.\ Kiefer is with The European Institute for Molecular Imaging, Münster, Germany.\\
  Corresponding author: Xiaoyi Jiang (xjiang@uni-muenster.de)
}
\let\thefootnote\svthefootnote

\begin{abstract}
Recent advances in 3D imaging technologies provide novel insights to researchers and reveal finer and more detail of examined specimen, especially in the biomedical domain, but also impose huge challenges regarding scalability for automated analysis algorithms due to rapidly increasing dataset sizes. In particular, existing research towards automated vessel network analysis does not consider memory requirements of proposed algorithms and often generates a large number of spurious branches for structures consisting of many voxels.
  Additionally, very often these algorithms have further restrictions such as the limitation to tree topologies or relying on the properties of specific image modalities.
  We present a scalable pipeline (in terms of computational cost, required main memory and robustness) that extracts an annotated abstract graph representation from the foreground segmentation of vessel networks of arbitrary topology and vessel shape.
  Only a single, dimensionless, a-priori determinable parameter is required.
  By careful engineering of individual pipeline stages and a novel iterative refinement scheme we are, for the first time, able to analyze the topology of volumes of roughly 1TB on commodity hardware.
  An implementation of the presented pipeline is publicly available in version 5.1 of the volume rendering and processing engine Voreen\footnote{\url{https://www.uni-muenster.de/Voreen/}}.
\end{abstract}


\IEEEpeerreviewmaketitle

\section{Introduction}
\begin{figure}[b]
    \includegraphics[width=\columnwidth]{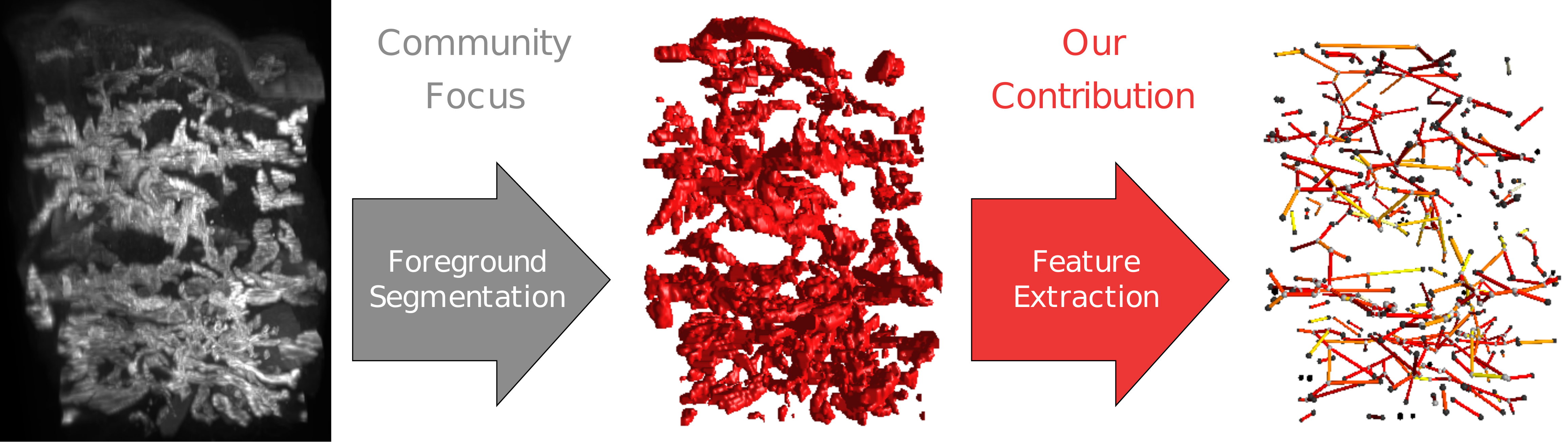}
    \caption{
      Typical pipeline for processing of 3D images of vascular networks.
      While the research community has mostly focused on the foreground segmentation task~(e.g., \cite{hu2017cerebral}), this paper presents a widely applicable analysis based on (potentially very large) binary volumes as input.
    }\label{fig:network_analysis_example}
\end{figure}

\IEEEPARstart{T}{he} study of vascular networks using modern 3D imaging techniques has become an increasingly popular topic of interest in biomedical research~\cite{hagerling2013novel, blinder2013cortical, bentley2014role, rene2017,Nazir2020,Ma2021}, as 2D slice analysis is limited to a small subset of the available data and cannot capture the 3D structure of vessel networks \cite{campinho2018three}.
At the same time, analysis of 3D datasets by visual inspection is error prone \cite{hamilton2009quantification}.
Thus, for quantifiable results or novel insights into data properties invisible to the human observer, automatic image processing techniques are required.
In light of different imaging techniques, modalities and problem domains, the generation of a voxel-wise foreground segmentation, can be seen as a sensible intermediate step for the automatic processing of vascular network images (\autoref{fig:network_analysis_example}).
However, the next step of calculating topological, morphological or geometric features -- a key requirement for the application in biomedical research and beyond -- has not received sufficient attention from the research community~\cite{zhao2019retinal}, which is mostly focused on the segmentation domain~\cite{lesage2009review,hu2017cerebral, moccia2018blood}.

In recent years, improvements in imaging technology and procedures are providing images of increasing resolution (in terms of the number of voxels per unit volume) and thus offer new chances for finer grained analysis, but also pose new challenges for image analysis algorithms.
The analysis of fine structures down to the level of capillary networks promises great insights.
At the same time it is desirable to analyze large networks at once in order to obtain as much topological information as possible and avoid artifacts at the boundary of the volume.
Modern microscopy hardware generates a single volumetric image file of hundreds of GBs or even TBs (the same order of magnitude as hard drive capacity) per sample which existing vessel analysis approaches are not able to process.
Due to rapid advances in imaging technology and data sizes, this problem extends from commodity hardware to specialized workstations.
Consequently, for a vessel network analysis pipeline to be useful now and in the future, it has to be \textit{scalable} both in terms of memory and runtime, and it should be invariant with regard to increases in image resolution.
For application in a wide range of biomedical domains, a pipeline should further not rely on specific imaging conditions, dataset-dependent parameters, a tubular vessel shape, isotropic image resolution or a specific network topology.

In this paper  we present a pipeline engineered to fulfill these requirements while extracting the topology, centerlines and edge associated features from binary volumetric images.
This is achieved by our \textit{main contribution}, a novel iterative refinement approach and careful design of all pipeline stages.


In the remainder of this paper we will first discuss related work and will then describe the proposed pipeline conceptually.
Next, we will elaborate the aforementioned requirements before describing individual pipeline stages in more detail with special attention to these requirements.
Finally, we will quantitatively evaluate different aspects of proposed pipeline with regard to the requirements and conclude our work.

\section{Related Work}\label{sec:related_work}
Chen et al.~\cite{chen2009generation} use the neighborhood-based voxel classification definitions of~\cite{lee1994building} for analysis of hepatic vasculature.
They also discuss other alternative voxel-based skeleton extraction methods, but note that they are not suitable:
Distance transformation-based methods~\cite{choi2003extraction} do not guarantee conntectedness of the extracted skeleton, while Voronoi-skeleton methods~\cite{ilg1992application} are comparatively time consuming.
Furthermore, Chen et al.~\cite{chen2009generation} perform extensive analysis to denote a single voxel of the skeleton as a branch point.
They extract a graph by following skeleton voxels in a tree search, breaking up cycles in the graph.
They do not perform any operations to remove erroneous branches -- likely because for the processed low resolution volumes the effect of noise is negligible.

Drechsler and Laura~\cite{drechsler2010hierarchical} extend~\cite{chen2009generation} by decomposing the generated skeleton into segments, computing a number of properties for each segment and generating a labeled graph structure from the segments.
The authors observe that their algorithm is very sensitive to noise in the foreground segmentation, but do not propose any means to reduce this effect.

In~\cite{chen2011thinning} Chen et al.\ present an alternative thinning-based skeletonization method for analysis of hepatic vasculature as part of the previously published pipeline~\cite{chen2009generation}.
They use the voxel classification methods of~\cite{lee1994building}, but present a different algorithm.
As a preprocessing step they suggest filling-hole techniques and morphological operations to remove cavities and irregularities on the surface of the object.

Chothani et al.~\cite{chothani2011automated} propose a pipeline for tracing of neurites from light microscopy image stacks.
It comprises foreground segmentation, skeletonization using a voxel coding algorithm~\cite{zhou99efficient}, tree construction using the scalar field of the skeletonization step and refinement including pruning based on length as well as separation of branches that appear to be connected due to limited image resolution.
The separation of branches in 3D should not be necessary using higher resolution images.
The authors note that in this case due to the increased volume size, advances in processing capabilities are required.

Almasi et al.~\cite{almasi2015novel} present a method of extracting graphs of microvascular networks from fluorescence microscopy image stacks.
In contrast to other methods operating on binary images they use imaging method specific information to improve the detection of vascular branches with disturbed image signal.

Cheng et al.~\cite{cheng2019detecting} analyze connected 3D structures such as vessel systems and metal foams in binary volumes.
They use topological thinning~\cite{couprie2007discrete} and merge resulting branching points using an algorithm based on regional maximal spheres and maximal inscribed spheres with regard to the object surface.
While their algorithm is linear in complexity, even for volumes of $10^9$ voxels (a gigabyte assuming one byte per voxel) it requires hours of computation time.

In~\cite{babin2018skeletonization}, the authors use a non-topology preserving distance field guided voxel thinning algorithm.
They construct a graph from an intermediate representation based on voxel neighborhood, but do not describe an efficient implementation.
The thinning step creates holes in irregularly shaped vessels which is used to detect arteriovenous malformations, but makes is unsuitable for irregularly shaped (e.g., lymphatic) vessels.

\textit{Rayburst sampling}~\cite{rodriguez2006rayburst} extracts precise measurements of radius, volume and surface of spatially complex histopathologic structures (given a centerline representation of a vessel).
The method operates on grey value images and detects the surface by sampling along rays originating from the measurement point.
For each point a large, but configurable number of rays is emitted to estimate the local morphology.

Finally, as many vessel network analysis pipelines include a skeletonization step, the work of Bates et al.~\cite{bates2017extracting} is worth of note.
The authors demonstrate binary vessel segmentation using convolutional recurrent neural networks, including a variant centerline extraction.
However, when compared to traditional skeletonization algorithms, the complex decision process of ANNs has disadvantages:
No hard guarantees of connectedness or the thickness of skeletal branches are available (without further processing).
Additionally, this approach may have problems with high resolution datasets where the vessels are wider than the field of view of the network, thus making it impossible to determine the centerline position.

All of the above works either do not mention processing of very large data samples (and thus likely do not support it) or explicitly mention that large datasets pose a problem for their method.
In this work we want to fill this gap by presenting a widely applicable pipeline that is engineered to handle very large input volumes and presented in the next section.

\section{Pipeline}\label{sec:pipeline}
\begin{figure}[t]
    \includegraphics[width=\columnwidth]{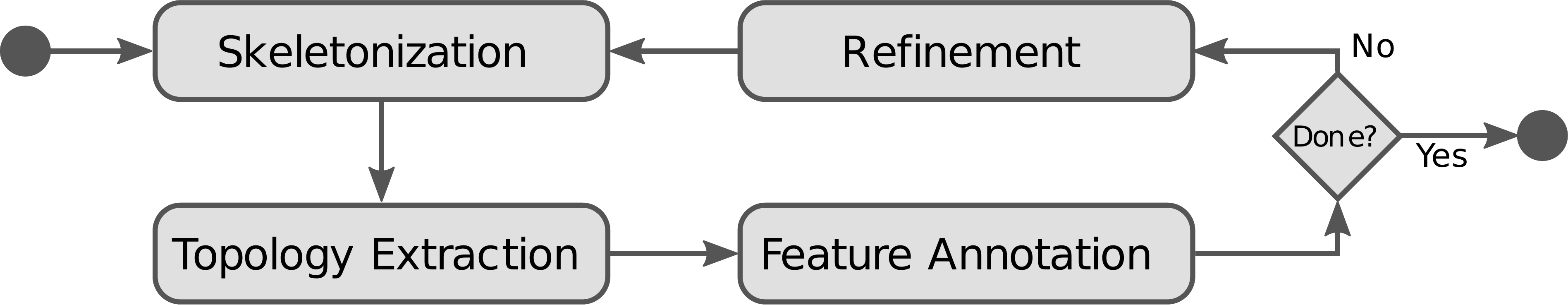}
    \caption{
      A schematic overview of the proposed pipeline.
    }\label{fig:pipeline}
\end{figure}
The proposed pipeline comprises four stages which are evaluated iteratively (see \autoref{fig:pipeline}).
In the first stage (\textit{Skeletonization}) the foreground of the binary input volume is reduced to a voxel skeleton using a topological thinning-based algorithm, similar to~\cite{drechsler2010hierarchical}.
We use a modified version of~\cite{lee1994building} which can be implemented efficiently both in terms of computational complexity and disk access to the out-of-core dataset.
Next, we build a graph representation from the voxel skeleton.
In contrast to \cite{drechsler2010hierarchical} this \textit{Topology Extraction} stage is implemented as a single sweep over volume, engineered for memory locality resulting in fewer disc accesses.
In the \textit{Feature Annotation} stage, we compute a set of morphological and geometrical properties for all edges of the graph.
This stage is subdivided into two steps.
First, the Voxel-Branch-Assignment determines for each foreground voxel which branch it is associated to.
The Feature Extraction step uses this mapping to efficiently calculate (geometrical~\cite{drechsler2010hierarchical} and additional morphological) features in single sweep over the volume.
Then, in the \textit{Refinement} stage, the graph is pruned using previously calculated features, using the scale invariant, dimensionless property bulge\_size.
Since pruning invalidates the Voxel-Branch-Assignment, the first three stages of the pipeline have to be reevaluated.
As shown in evaluation, this step is essential for application on high resolution images.
This Extraction-Refinement cycle is repeated until a fixed point is reached.
We want to stress that the novel iterative refinement approach is essential to obtain meaningful results from very large datasets, as demonstrated in~\autoref{sec:evaluation}.

\section{Requirements for Broad Biomedical Application}\label{sec:requirements}
As discussed before, the proposed pipeline, in contrast to previous work, is engineered to fulfill a set of requirements that are essential for a wide range of biomedical applications.
For the discussion in the remainder of this paper, in this section we will structure, motivate and expand on the requirements briefly mentioned in the introduction.

\textbf{Primary Requirements:}
The method should be scalable with regard to \textbf{runtime}~(P1).
As all voxels of a volume can be expected to be taken into account by the method, the runtime can be expected to be at least linear in the number of voxels $n$.
In order to be applicable to increasingly large datasets, a runtime of $\O(n^2)$ would be unacceptable, but quasilinear algorithms (e.g., in $\O(n\log n)$) can still be executed.
Beyond computational complexity, special care has to be taken for example in terms of memory access locality to avoid an increase in computation time by a large constant factor.

The method should be scalable with regard to \textbf{memory requirements}~(P2).
We can expect the input datasets to fit onto (commodity) non-volatile storage (disk), but not necessarily in main memory.
While the price per capacity of both disks and main memory fall exponentially, the rate of (exponential) decrease is higher for disks than for memory~\cite{grochowski1998emerging,historical2019mccallum}.
Consequently, the ratio of average disk size to average RAM size grows over time.
The exact relationship is not easy to establish as it changes over time~\cite{historical2019mccallum}.
Here we assume that disk and volatile memory size \textit{roughly} grow in a relationship of $m$ to $m^{\frac{2}{3}}$ for a disk size of $m$.
Therefore the required main memory size of the method should not exceed $\O(m^{\frac{2}{3}})$.

The method should exhibit \textbf{invariance with regard to image resolution}~(P3).
For a fixed size specimen, an increase in image resolution results in a dramatic increase in surface noise related artifacts for simple topological thinning-based methods which is unacceptable for analysis of large datasets.

\textbf{Secondary Requirements:} From the desire to apply the method in practice we derive a set of further requirements that are not directly related to scalability:

The method should be \textbf{unbiased} (S1), i.e., not dependent on a set of parameters that are required to be selected carefully and \textit{correctly} depending on the input image.

The method should be \textbf{robust in terms of deviations from the tube shape} (S2) of the analyzed network structure.
While blood vessels are tubular, lymphatic vasculature as an example is highly irregular, but should still be processed correctly.

The method should handle volumes of \textbf{anisotropic resolution} (S3).
Volumetric imaging techniques often create datasets with voxel spacing that differs between coordinate axes.
As an example, in light sheet microscopy datasets are constructed from a series of (2D) slices for which the spacing is independent from the x-y-resolution.
Methods operating on the voxel grid of a volume have to consider this.

The method should be able to fully analyze networks of \textbf{arbitrary topology} (S4).
Existing methods often assume a tree structure and either fail to operate on other topologies or drop information~\cite{chen2009generation}.
However, for lymphatic vessels and capillaries, or even larger structures (\textit{cerebral arterial circle}) the assumption of a tree topology is invalid.

The method should be able to analyze images \textbf{independently of the imaging conditions} (S5).
Concrete imaging conditions (i.e., distribution of gray values or fluorescent staining techniques in microscopy imaging) vary between domains.
In order to be widely applicable, the method should not be dependent on concrete models of imaging conditions.
Our method achieves this by operating solely on binary input data and leveraging the broad range of other research on vessel image segmentation~\cite{lesage2009review,moccia2018blood}.

\section{Method Details}\label{sec:method}

\begin{figure}[t]
    \includegraphics[width=\columnwidth]{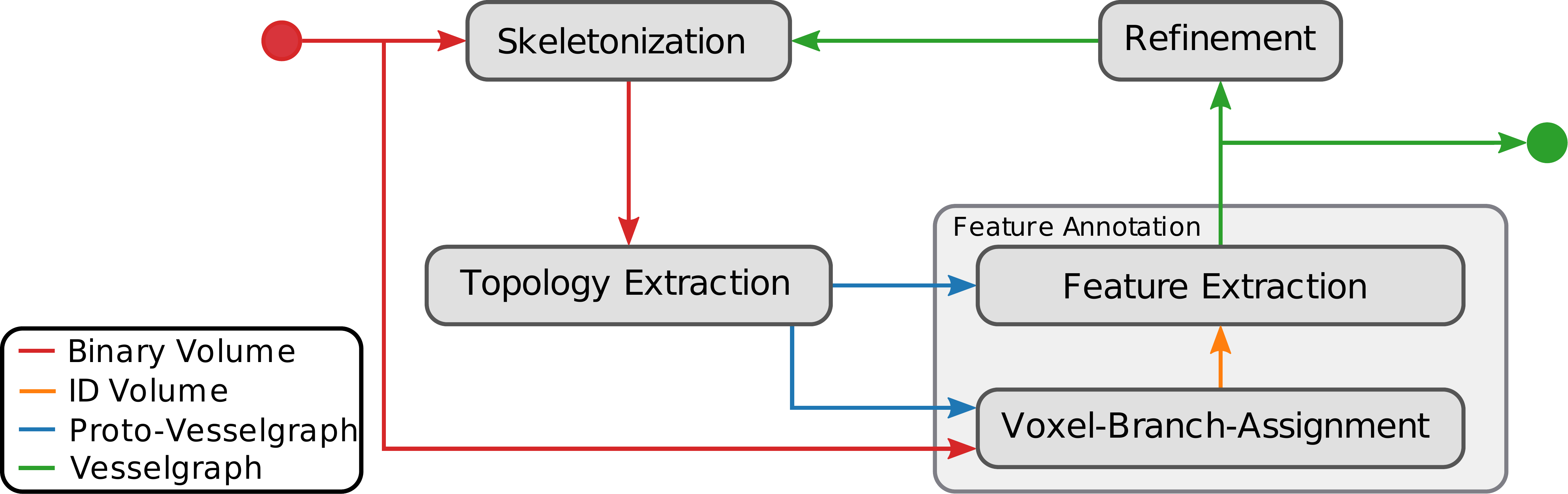}
    \caption{
      A schematic overview of the data flow in the proposed pipeline.
      Dashed lines signify an optional dependency on the data for the particular stage.
      For skeletonization topological information is not available in the first iteration of the pipeline.
    }\label{fig:method_overview}
\end{figure}

In this section, we will present the algorithmic details of the proposed pipeline with special attention to the requirements for broad biomedical application.
\autoref{fig:method_overview} provides an overview of the data flow between individual stages of the pipeline.

\subsection{Skeletonization}\label{sec:skeletonization}
Like other vessel network analysis pipelines \cite{chen2009generation,drechsler2010hierarchical}, we use a modified version of the established topological thinning algorithm by Lee et al.~\cite{lee1994building} which has the advantage that any voxel can be evaluated in terms of these properties by only considering its 26-Neighborhood, which is advantageous when operating on very large volumes.
Furthermore, we are not aware that other approaches solve the inherent problems of analysis of large datasets.
In the original formulation, the algorithm iteratively removes voxels that lie on the surface and whose deletion does not change the topology of the object in 6 (static) subiterations until no more voxels can be removed.

For an efficient implementation, further considerations are necessary.
We model the surface of the object explicitly as a sequence of voxel positions, which is initialized by scanning the volume before the first iteration step.
In subsequent subiterations, we build the \textit{active surface} (voxels that can potentially be deleted in the next iteration) by retaining voxels from the previous active surface that are not considered during the current subiteration and adding all foreground voxels in the 26-Neighborhood of a voxel after its deletion.
If a voxel was considered, but not deleted during the current subiteration, it will be removed from the active surface even although it is still part of the surface of the object.
If one of its neighbors is deleted, it will be re-added to the active surface and reconsidered for deletion in the following iteration.
This implementation has a runtime of $\O(n)$ in the number of voxels in the volume~(P1):
In each subiteration, only voxels in the active surface are considered.
As a voxel is either deleted entirely or removed from the active surface after each iteration (i.e., 6 subiterations) and only added again once one of its 26 neighbors is deleted, it will be considered for deletion a constant amount of times.
In order to fulfill requirement P2, the volume is stored on disk and dynamically mapped to memory using operating system capabilities.
Constant runtime improvements could be observed using a compressed (2 bits per voxel) representation and storing the volume as 32x32x32 (8 KB) blocks in linear memory, thus reducing disk access by exploiting the spatial locality of surface voxels.
Active surfaces are stored as on disk (linear) voxel positions in sorted sequences.
During a subiteration the previous active surface is read from front to back.
Simultaneously, the new active surface is built slice-by-slice in memory by collecting positions, and sorting and removing duplicates before writing to disk, requiring $\O(m^\frac{2}{3})$ main memory~(P2).

To account for volumes with anisotropic resolution, we keep track of the real-world depth of voxel layers deleted for each direction and select the direction with the lowest total depth as the direction for the next subiteration.
This way, the \textit{speed} of voxel deletion is equalized on average for highly anisotropic volumes (S3).
If the last subiterations of all 6 directions (which may have happened out of order) did not delete any voxels, the active surface is empty and the skeletonization is completed.

\subsection{Topology Extraction}\label{sec:topology_extraction}
Previous work~\cite{chen2009generation,babin2018skeletonization} does not describe an efficient implementation of topology extraction for large input datasets.
For example, Chen et al.~\cite{chen2009generation} perform a tree search following skeleton voxels through the volume.
This is unsuitable for large volumes as it either requires keeping the complete volume in memory or involves very frequent random access to hard disk with high latency.
Additionally, their method breaks loops in the graph and always creates a tree topology.

In constrast, the topology extraction in the proposed pipeline extracts the complete centerline graph in a \textit{single} pass over the volume using a modified version of the streaming connected component finding algorithm by Isenburg and Shewchuk~\cite{isenburg2009streaming}.
Instead of all foreground components, we consider the three skeleton voxels classes (regular (2 neighbors), end ($<2$ neighbors), and branch ($>2$ neighbors)~\cite{chen2009generation}) separately and extract components for each.
Individual connected components of end or branch voxels make up nodes in the graph.
Connected components of regular voxels form lines within the volume and have (except for the separately handled case of closed loops) exactly two end points.
Unlike~\cite{chen2009generation}, we do not force the position of nodes to be defined by a single voxel and instead use the barycenter of the connected region of voxels.
Using the two end points of a segment of regular voxels we can find the nodes that are connected by the path of regular voxels and thus construct an edge.
To perform this node-edge-matching process efficiently for large graphs, for each node voxel (i.e., an end or branch voxel) we insert a reference to the original node into a (static, on-disk) k-d tree.
Thus, for each run of regular voxels we can efficiently query the position of node voxels that are 26-neighbors of its tips, and thus link node and edge data structures.
Extracted nodes and edges (including centerlines) form the \textit{Proto-Vesselgraph}, are each stored in files on the disk and are dynamically mapped to memory.
Furthermore, the algorithm of~\cite{isenburg2009streaming} (and this modification) can be shown to only require $\O(m^\frac{2}{3})$ main memory~(P2).
Moreover, we extract the whole graph in a single pass over the volume in $\O(n \log n)$ time~(P1) and do not make any assumptions about the topology of the extracted network (S4).

Since the individual branch voxels necessarily lie on voxel positions of the original volume, they resemble a ragged centerline.
This both artificially increases the line length and produces a larger number of ambiguous cases in the Voxel-Branch-Assignment.
We therefore smooth all centerlines of the Proto-Vesselgraph using local bezier curves.

\subsection{Voxel-Branch Assignment}\label{sec:voxel_branch_assignment}
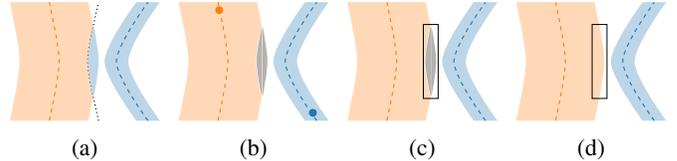
\begin{figure}
  \definecolor{leftcolor}{RGB}{255,127,14}
  \definecolor{rightcolor}{RGB}{31,119,180}
  \def\top{3.0}
  \def\bottom{0.0}
  \def\middle{1.5}

  \def\fatouter{1.0}
  \def\fatinner{1.25}
  \def\fatradius{1.0}

  \def\thinouter{3.5}
  \def\thininner{2.7}
  \def\thinradius{0.3}

  \def\intersectiontop{3.25*0.75}
  \def\intersectionbottom{0.75*0.75}
  \def\subfigboxwidth{0.49}

  \def\fillopacity{30}

  \ADD{\thininner}{\fatinner}{\innersum}
  \DIVIDE{\innersum}{2}{\intersectioninner}

  \ADD{\thinouter}{\fatouter}{\outersum}
  \DIVIDE{\outersum}{2}{\intersectionouter}

  \begin{subfigure}[t]{0.24\columnwidth}
    \resizebox{\columnwidth}{!}{
      \begin{tikzpicture}
        \fill[rightcolor!\fillopacity]
        plot [smooth] coordinates {(\thinouter-\thinradius,\bottom) (\thininner-\thinradius,\middle) (\thinouter-\thinradius,\top)}
        --
        plot [smooth] coordinates {(\thinouter+\thinradius,\top) (\thininner+\thinradius,\middle) (\thinouter+\thinradius,\bottom)}
        ;

        \fill[leftcolor!\fillopacity]
        plot [smooth] coordinates {(\fatouter-\fatradius,\bottom) (\fatinner-\fatradius,\middle) (\fatouter-\fatradius,\top)}
        --
        plot [smooth] coordinates {(\fatouter+\fatradius,\top) (\fatinner+\fatradius,\middle) (\fatouter+\fatradius,\bottom)}
        ;

        \draw [dashed, leftcolor, thick] plot [smooth] coordinates {(\fatouter,\bottom) (\fatinner,\middle) (\fatouter,\top)};
        \draw [black!0, thick, name path = FATWALLRIGHT]        plot [smooth] coordinates {(\fatouter+\fatradius,\bottom) (\fatinner+\fatradius,\middle) (\fatouter+\fatradius,\top)};

        \draw [dashed, rightcolor, thick]  plot [smooth] coordinates {(\thinouter,\bottom) (\thininner,\middle) (\thinouter,\top)};

        \draw [black, dotted, thick, name path = VORONOI] plot [smooth] coordinates {(\intersectionouter,\bottom) (\intersectioninner,\middle) (\intersectionouter,\top)};

        \fill [rightcolor!\fillopacity, intersection segments={ of=FATWALLRIGHT and VORONOI, sequence={L2--R2} }];

      \end{tikzpicture}
    }
    \caption{
    }\label{fig:cut_off_regions_a}
  \end{subfigure}
  \begin{subfigure}[t]{0.24\columnwidth}
    \resizebox{\columnwidth}{!}{
      \begin{tikzpicture}
        \fill[rightcolor!\fillopacity]
        plot [smooth] coordinates {(\thinouter-\thinradius,\bottom) (\thininner-\thinradius,\middle) (\thinouter-\thinradius,\top)}
        --
        plot [smooth] coordinates {(\thinouter+\thinradius,\top) (\thininner+\thinradius,\middle) (\thinouter+\thinradius,\bottom)}
        ;

        \fill[leftcolor!\fillopacity]
        plot [smooth] coordinates {(\fatouter-\fatradius,\bottom) (\fatinner-\fatradius,\middle) (\fatouter-\fatradius,\top)}
        --
        plot [smooth] coordinates {(\fatouter+\fatradius,\top) (\fatinner+\fatradius,\middle) (\fatouter+\fatradius,\bottom)}
        ;

        \draw [dashed, leftcolor, thick] plot [smooth] coordinates {(\fatouter,\bottom) (\fatinner,\middle) (\fatouter,\top)};
        \fill [leftcolor] (\fatouter+0.03,\top-0.2) circle[radius=0.1];

        \draw [black!0, thick, name path = FATWALLRIGHT]        plot [smooth] coordinates {(\fatouter+\fatradius,\bottom) (\fatinner+\fatradius,\middle) (\fatouter+\fatradius,\top)};

        \draw [dashed, rightcolor, thick]  plot [smooth] coordinates {(\thinouter,\bottom) (\thininner,\middle) (\thinouter,\top)};
        \fill [rightcolor] (\thinouter-0.1,\bottom+0.2) circle[radius=0.1];

        \draw [transparent!0, name path = VORONOI] plot [smooth] coordinates {(\intersectionouter,\bottom) (\intersectioninner,\middle) (\intersectionouter,\top)};

        \fill [black!\fillopacity, intersection segments={ of=FATWALLRIGHT and VORONOI, sequence={L2--R2} }];

      \end{tikzpicture}
    }
    \caption{
    }\label{fig:cut_off_regions_b}
  \end{subfigure}
  \begin{subfigure}[t]{0.24\columnwidth}
    \resizebox{\columnwidth}{!}{
      \begin{tikzpicture}
        \fill[rightcolor!\fillopacity]
        plot [smooth] coordinates {(\thinouter-\thinradius,\bottom) (\thininner-\thinradius,\middle) (\thinouter-\thinradius,\top)}
        --
        plot [smooth] coordinates {(\thinouter+\thinradius,\top) (\thininner+\thinradius,\middle) (\thinouter+\thinradius,\bottom)}
        ;

        \fill[leftcolor!\fillopacity]
        plot [smooth] coordinates {(\fatouter-\fatradius,\bottom) (\fatinner-\fatradius,\middle) (\fatouter-\fatradius,\top)}
        --
        plot [smooth] coordinates {(\fatouter+\fatradius,\top) (\fatinner+\fatradius,\middle) (\fatouter+\fatradius,\bottom)}
        ;

        \draw [dashed, leftcolor, thick] plot [smooth] coordinates {(\fatouter,\bottom) (\fatinner,\middle) (\fatouter,\top)};

        \draw [black!0, thick, name path = FATWALLRIGHT]        plot [smooth] coordinates {(\fatouter+\fatradius,\bottom) (\fatinner+\fatradius,\middle) (\fatouter+\fatradius,\top)};

        \draw [dashed, rightcolor, thick]  plot [smooth] coordinates {(\thinouter,\bottom) (\thininner,\middle) (\thinouter,\top)};

        \draw [transparent!0, name path = VORONOI] plot [smooth] coordinates {(\intersectionouter,\bottom) (\intersectioninner,\middle) (\intersectionouter,\top)};

        \fill [black!\fillopacity, intersection segments={ of=FATWALLRIGHT and VORONOI, sequence={L2--R2} }];

        \draw[draw=black] (\intersectioninner-0.05,\intersectiontop) rectangle (\intersectionouter+0.05,\intersectionbottom);

      \end{tikzpicture}
    }
    \caption{
    }\label{fig:cut_off_regions_c}
  \end{subfigure}
  \begin{subfigure}[t]{0.24\columnwidth}
    \resizebox{\columnwidth}{!}{
      \begin{tikzpicture}
        \fill[rightcolor!\fillopacity]
        plot [smooth] coordinates {(\thinouter-\thinradius,\bottom) (\thininner-\thinradius,\middle) (\thinouter-\thinradius,\top)}
        --
        plot [smooth] coordinates {(\thinouter+\thinradius,\top) (\thininner+\thinradius,\middle) (\thinouter+\thinradius,\bottom)}
        ;

        \fill[leftcolor!\fillopacity]
        plot [smooth] coordinates {(\fatouter-\fatradius,\bottom) (\fatinner-\fatradius,\middle) (\fatouter-\fatradius,\top)}
        --
        plot [smooth] coordinates {(\fatouter+\fatradius,\top) (\fatinner+\fatradius,\middle) (\fatouter+\fatradius,\bottom)}
        ;

        \draw [dashed, leftcolor, thick] plot [smooth] coordinates {(\fatouter,\bottom) (\fatinner,\middle) (\fatouter,\top)};

        \draw [black!0, thick, name path = FATWALLRIGHT]        plot [smooth] coordinates {(\fatouter+\fatradius,\bottom) (\fatinner+\fatradius,\middle) (\fatouter+\fatradius,\top)};

        \draw [dashed, rightcolor, thick]  plot [smooth] coordinates {(\thinouter,\bottom) (\thininner,\middle) (\thinouter,\top)};

        \draw[draw=black] (\intersectioninner-0.05,\intersectiontop) rectangle (\intersectionouter+0.05,\intersectionbottom);

      \end{tikzpicture}
    }
    \caption{
    }\label{fig:cut_off_regions_d}
  \end{subfigure}
  \caption{
    If the centerline of a small vessel (blue) is closer to the surface of a larger vessel (yellow) than its own centerline, some voxels of the larger vessel will be assigned to the smaller vessel (a).
    After the initial Voronoi Mapping, connected foreground components of the same ID are remapped using IDs from seed points of branches (b).
    Unlabeled regions are identified (c) and flood-filled from labeled voxels (d).
  }\label{fig:cut_off_regions}
\end{figure}

Drechsler and Laura~\cite{drechsler2010hierarchical} calculate the $volume$ property of edges by assigning voxels of the foreground segmentation to the nearest centerline point.
As illustrated in \autoref{fig:cut_off_regions_a}, this creates incorrect results in some cases.
While these errors are potentially not as severe for the calculated $volume$, morphological properties based on the radius of the vessel can be heavily affected.
In order to correctly calculate edge-associated properties for the previously created Proto-Vesselgraph structure, we therefore first create a mapping between voxels of the foreground segmentation and the edges of the Proto-Vesselgraph.
This process is performed in 4 steps that are described below and illustrated in \autoref{fig:cut_off_regions}.

\subsubsection{Centerline Voronoi Mapping}
As a first approximation of the edge ID map, we find the nearest centerline point for each voxel of the foreground segmentation.
We accelerate the search using a static on-disk k-d tree of all centerline points of all edges of the Proto-Vesselgraph, each annotated with the corresponding edge ID\@.
As we iterate over all $n$ foreground voxels and perform a $\log n $ search in the k-d tree for each, the total runtime is within $\O(n \log n)$, as is the construction of the k-d tree~(P1).
We iterate over the whole volume in 32x32x32 blocks of voxels for spatial locality of lookups, thus reducing the need to randomly reload parts of the k-d tree from disk.

\subsubsection{Connected Component Remapping}
As illustrated in \autoref{fig:cut_off_regions_a}, matching voxels via minimal euclidean distance does not yield a proper voxel wise vessel segment map in all cases.
When two vessels with highly dissimilar radii lie close to each other, some voxels of the larger vessel will be ascribed to the smaller vessel.
These cut off regions have to be identified and remapped.
We perform a modified connected component analysis~\cite{isenburg2009streaming} on the generated edge ID volume in which we consider two voxels \textit{mergable} if they have the same ID.
Then we map the connected component IDs back to the edge IDs using a table constructed by sampling at one centerline point of each edge in the result of the connected component analysis.

\subsubsection{Cut-off Region Identification}
Cut-off regions do not have an associated edge ID and can therefore be identified using another pass of a connected component analysis~\cite{isenburg2009streaming} where only voxels with undefined IDs are considered.
In a single pass over the whole volume, we collect the axis aligned bounding boxes for each cut-off region (see \autoref{fig:cut_off_regions_c}).

\subsubsection{Cut-off Region Flooding}
Previously identified cut-off regions are relabeled by flooding all voxels without edge IDs starting from adjacent voxels with valid edge IDs.
For the simple case of a region cut-off from a single vessel section, this fills the region with the section's edge ID\@.
In a more complicated case of a cut-off region adjacent to multiple different vessel sections, the resulting labeling approximates the $L1$-Voronoi-cells spanned from the surface of adjacent regions.
A side effect of this procedure is that foreground regions that do not resemble (complete) vessels at the boundary of the volume (and thus do not contain any centerline voxels), are \textit{not} flooded with valid IDs in this step and ignored in subsequent steps.

For efficiency, all identified disconnected regions are processed separately as cuboidal subvolumes and collected during a single pass over the ID-volume.
Currently valid bounding boxes of cut-off regions are organized in interval trees.
For the z-axis, a single tree references all bounding boxes.
For each slice, the active regions can be queried to construct an interval tree for the y-axis.
For each line in the slice, using the y-axis tree an interval tree for the x-axis can be constructed.
For each voxel in the line, all active regions can be queried from the x-axis tree so that the current voxel value can be written to the corresponding subvolumes.
After this collection step, each subvolume is flooded separately.
Similar to the skeletonization step, we use the concepts of an active surface, slice-wise processing and memory mapping of files on disk to guarantee linear runtime~(P1) and $\O(m^\frac{2}{3})$ memory requirement~(P2).

The results are written back to the ID volume in a process similar to the collection of the subvolumes.
The number of disconnected regions is bounded by $n$.
Consequently, the construction of the range trees and the querying for each voxel are within $O(n\log n)$~(P1).
All subvolumes and range trees can be stored on disk and mapped dynamically to memory and therefore do not require additional memory~(P2).

\subsection{Feature Extraction}\label{sec:feature_extraction}
Using the Proto-Vesselgraph and the generated edge ID volume, we can compute the same edge properties as Drechsler and Laura~\cite{drechsler2010hierarchical}.
Some geometric features can be computed directly from the Proto-Vesselgraph.
These include $length$ (arc-length of the centerline), $distance$ (euclidean distance of connected nodes) and straightness ($\frac{distance}{length}$).
Drechsler and Laura~\cite{drechsler2010hierarchical} propose \textit{curveness} ($\frac{length}{distance}$), but as the $distance$ of any vessel segment is guaranteed to be smaller than its arc-$length$, we argue that straightness (with a guaranteed domain of $[0,1]$) is superior to curveness ($[1, \infty)$).

For other features we collect information from the edge ID volume and the Proto-Vesselgraph.
For each foreground voxel in the edge ID volume, we find the corresponding edge and query the closest of its centerline point(s).
For efficient lookups, the centerline points are organized in a (static, on-disk) k-d tree for each edge.
We equally distribute and store the volume occupied by the foreground voxels alongside the corresponding centerline point(s).
Surface voxels are assigned to the closest centerline point, for which thus the minimum, maximum and average surface distance is computed.

Values collected for individual centerline points can be aggregated to compute more per-edge features:
For each of the surface distance-based values ($minimum$, $maximum$, $average$, $roundness := \frac{minimum}{maximum}$) we compute the mean and standard deviation, totalling in 8 \textit{additional} morphological features not discussed in~\cite{drechsler2010hierarchical}.
Compared to~\cite{rodriguez2006rayburst} this provides a relatively coarse understanding of the local vessel morphology, but is a good approximation that can be computed efficiently.
Finally, the per-voxel volume can be accumulated to compute the total $volume$ occupied by each vessel segment in the vessel graph.
The average cross-section of a vessel is obtained by dividing the volume by the length~\cite{drechsler2010hierarchical}.

In total, the number of centerline points is obviously bounded by $n$ so that the cost of building and querying the k-d trees is within $\O(n\log n)$.
Additionally, the label volume is accessed in 32x32x32 chunks improving memory locality of searches in the k-d trees~(P1).
All components of the Proto-Vesselgraph, the k-d trees and the created graph are stored on disk so that main memory requirements are met~(P2).

\subsection{Refinement}\label{sec:refinement}
Skeletonization algorithms tend to produce a number of small spurious branches, especially for vessels with irregular surfaces.
We therefore propose to refine the generated vessel graph by pruning spurious branches.
A simple, but scale-dependent way of defining \textit{deletability} is to set a global minimum length~\cite{chothani2011automated}.
However, this is problematic if vessels of different scales are present in a single dataset.
Instead, we propose \textit{bulge size} as a scale-independent dimensionless measure (S1).
Intuitively, the bulge size measures how far a bump, bulge or branch has to extend from a parent vessel in order to be considered a separate vessel.
This size is expressed relative to the radius of its parent vessel and itself, making it scale-independent.
More formally, the \textit{bulge size} is an edge feature, that is computed during the feature extraction \ref{sec:feature_extraction}, and is only defined for \textit{bulging edges}, i.e., edges that connect a leaf node (degree 1) and a branching point (degree $> 2$).

For all centerline points of the edge, we decide whether they are within a branching region (\textit{inner points}) or not (\textit{outer points}) during the feature extraction.
If a point has associated surface points, neither of which are adjacent to surface voxels of another edge, it is condidered an outer point.
An inner point is characterized by either not having any associated surface voxels or by having a surface voxel that is a 26-neighbor of a point that belongs to another edge.
The inner length of a node is the arc length defined by a run of centerline points starting from that node and ending at the last inner voxel.
This corresponds to the length of the piece of centerline that lies within a branching area.
The \textit{tip radius} is defined as the minimum distance to the surface measured from the centerline point closest to the node.
Without loss of generality, let $n_b$ and $n_e$ be the branching point node and leaf node, respectively.
\begin{align*}
  &bulge\_size(e=(n_b,n_e))\\= &\frac{length(e)-inner\_length(n_b)+tip\_radius(n_e)}{avgRadiusMean(e)}
\end{align*}
This definition provides a dimensionless measure of shape that performs well even for corner cases and can be computed efficiently during the feature extraction stage of the pipeline.
The calculation and applicability is illustrated in \autoref{fig:bulge_size}.

\definecolor{radiuscolor}{RGB}{255,127,14}
\definecolor{lengthcolor}{RGB}{31,119,180}
\definecolor{innerlengthcolor}{RGB}{44, 160, 44}
\definecolor{tipradiuscolor}{RGB}{216, 39, 40}

\newcommand\vesselbulgegraphic[1]{
  \def\top{#1}
  \def\tipdiff{0.3}
  \def\radius{0.67}
  \def\radiuslabely{\top+0.2}
  \def\maxbranchradius{1}
  \SUBTRACT{\top}{\tipdiff}{\tipy}
  \MIN{0}{\tipy}{\innerbordery}
    \begin{tikzpicture}

      \coordinate (tl) at (-2,0);
      \coordinate (tr) at (2,0);
      \coordinate (bl) at (-2,-1);
      \coordinate (br) at (2,-1);
      \coordinate (bulge) at (0, \top);
      \coordinate (branch) at (0,-0.5);
      \coordinate (leftend) at (-2,-0.5);
      \coordinate (rightend) at (2,-0.5);
      \coordinate (tip) at (0,\tipy);
      \coordinate (innerborder) at (0,\innerbordery);

      \fill[gray!20] (br) -- (bl) -- plot [smooth] coordinates {(tl) (-\maxbranchradius,0) (bulge) (\maxbranchradius,0) (tr)} -- (br);
      \draw [gray, thick] plot [smooth] coordinates {(tl) (-\maxbranchradius,0) (bulge) (\maxbranchradius,0) (tr)};
      \draw [gray, thick] plot coordinates {(br) (bl)};

      \draw [black, thick] plot coordinates {(leftend) (rightend)};
      \draw [black, thick] plot coordinates {(branch) (tip)};
      \fill (branch) circle[radius=0.1];
      \fill (tip) circle[radius=0.1];

      \path[|-|, lengthcolor, thick, transform canvas={xshift = 0.4cm}] (branch) edge (tip);

      \path[|-|, innerlengthcolor, thick, transform canvas={xshift = -0.2cm}] (branch) edge (innerborder);
      \path[dashed, innerlengthcolor, thick] (-\maxbranchradius,0) edge (\maxbranchradius,0);

      \path[|-|, tipradiuscolor, thick] (tip) edge (bulge);

      \path[|-|, radiuscolor, thick] (0, \radiuslabely) edge (\radius, \radiuslabely);
      \path[dashed, radiuscolor, thick] (\radius,0) edge (\radius,\tipy);

    \end{tikzpicture}
}

\begin{figure}[t]
  \begin{subfigure}[t]{0.49\columnwidth}
    \resizebox{\columnwidth}{!}{\vesselbulgegraphic{0.4}}
    \caption{$bulge\_size \approx \frac{1}{2}$}
  \end{subfigure}
  \begin{subfigure}[t]{0.49\columnwidth}
    \resizebox{\columnwidth}{!}{\vesselbulgegraphic{0.7}}
    \caption{$bulge\_size \approx 1$}
  \end{subfigure}

  \begin{subfigure}[t]{0.49\columnwidth}
    \resizebox{\columnwidth}{!}{\vesselbulgegraphic{1.5}}
    \caption{$bulge\_size \approx 2$}
  \end{subfigure}
  \begin{subfigure}[t]{0.49\columnwidth}
    \centering
      \resizebox{0.7\columnwidth}{!}{
        \begin{tikzpicture}
          \draw[black, thick, loop] (-0.2,1.8) -- (3.5,1.8) -- (3.5,-0.3) -- (-0.2,-0.3) -- (-0.2,1.8);
          \draw[|-|, radiuscolor, thick] (0,1.5) -- (0.5,1.5) node [right, black] {$avgRadiusMean$};
          \draw[|-|, lengthcolor, thick] (0,1.0) -- (0.5,1.0) node [right, black] {$length$};
          \draw[|-|, innerlengthcolor, thick] (0,0.5) -- (0.5,0.5) node [right, black] {$inner\_length$};
          \draw[|-|, tipradiuscolor, thick] (0,0.0) -- (0.5,0.0) node [right, black] {$tip\_radius$};
        \end{tikzpicture}
      }
    \caption{legend}
  \end{subfigure}
  \caption{
    Schematic depiction of the calculation of the \textit{bulge size} feature of an edge for three examples.
  }\label{fig:bulge_size}
\end{figure}
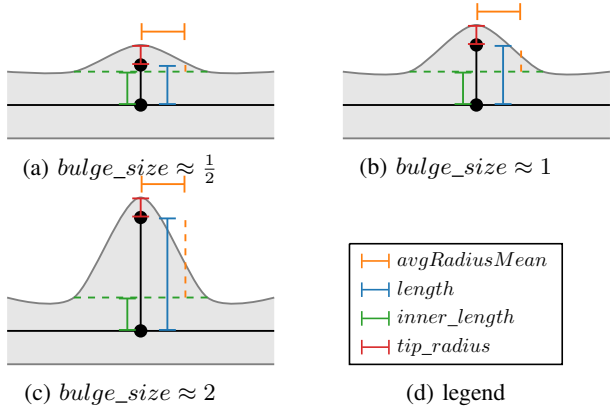

The pruning of spurious branches is a node-oriented, iterative approach outlined in \autoref{alg:graph_refinement}.
A pruning step iterates over all nodes in the graph and collects deletable branches.
All deletable edges and nodes that were only connected to those edges are then removed from the graph.
Additionally, two edges that share a common node of degree 2 are merged by connecting their centerlines and recomputing properties from the attributes of points in the combined centerlines.
The pruning step is repeated until a fixed point is reached.

\begin{algorithm}[t]
  \caption{Graph refinement algorithm}\label{alg:graph_refinement}
\begin{algorithmic}[1]
  \Function{refine}{$G, t$}
    \Repeat
    \State $D$ = \{\}
    \For {node $n \in G$}
      \State $E$ = edges connected to $n$
      \State $P$ = \{\}
      \For {$e \in E$}
        \If {$bulging(e) \land bulge\_size(e)<t$}
          \State $P = P \cup \{e\}$
        \EndIf
      \EndFor
      \While {$|E| - |P| < 2$} \Comment Retain two edges
        \State $P = P - \{\underset{e \in P}{\argmax} \, bulge\_size(e)\}$
      \EndWhile
      \State $D = D \cup P$
    \EndFor
    \State Delete edges in $D$ and remove orphaned nodes
    \State Remove nodes with degree 2 and merge edges
    \Until $D$ = \{\} \Comment Iterate until reaching fixed point
    \State \Return G
  \EndFunction
\end{algorithmic}
\end{algorithm}

While the refined \textit{graph} does not retain any signs of deleted edges, the simultaneously computed centerlines and edge properties are not preserved by the refinement procedure.
This happens because the volume regions associated with now deleted branches do not contribute to the properties of the remaining branches until the ID map and features are recomputed.
While this affects the bulge size, the pruning scheme will not erroneously delete branches, since the missing regions (lowered radii) cause the bulge size to be \textit{overrestimated}.

\subsection{Iterative Scheme}\label{sec:iterative_scheme}
In order to ensure that centerlines and edge properties match the refined graph, we employ an iterative scheme where the previously extracted and refined graph is fed back into the first stage of the pipeline to improve the generated results.

We modify the skeletonization algorithm to force it to generate a voxel skeleton with a topology that matches the refined graph of the previous iteration.
All voxels of nodes with degree 1 in the graph are set to be \textit{fixed} foreground during the skeletonization of the input volume.
This means that they are \textit{never} considered for deletion.
These voxels mark the beginning/end points of lines to be extracted during the skeletonization.
The skeletonization algorithm is modified to not preserve voxels at the end of voxel lines (non-\textit{line voxels} \cite{lee1994building}).
The resulting voxel skeleton connects all previously extracted endvoxels with medially positioned lines.

The iteration can be stopped if it reaches a fixed point.
This is the case if two consecutive iterations generate the same graph.
As the number of edges never increases, the check can be reduced to a simple integer comparison.
Optionally, an upper limit for the number of iterations can be specified.

\section{Evaluation and Discussion}\label{sec:evaluation}

In this section we evaluate and discuss the proposed pipeline in terms of the primary and secondary requirements for biomedical application (\autoref{sec:requirements}). All experiments were conducted on a consumer grade PC (AMD Ryzen 7 2700X (3,7 GHz), 32GB RAM, and 1 TB hard disk (Samsung NVMe SSD 960 EVO)).

\subsection{Runtime Scalability}
\begin{figure}[t]
  \begin{subfigure}[t]{0.32\columnwidth}
    \centering\includegraphics[width=0.7\textwidth]{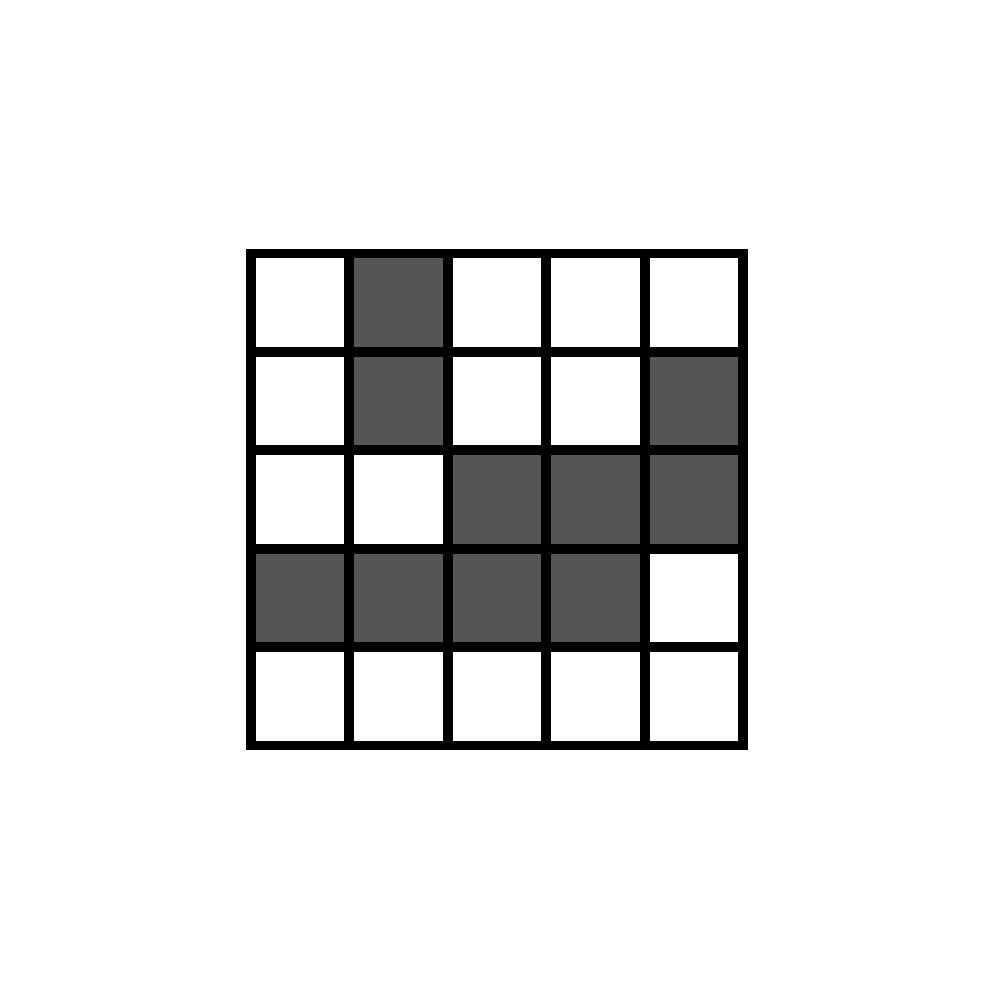}
    \caption{Original Image}
  \end{subfigure}
  \begin{subfigure}[t]{0.32\columnwidth}
    \centering\includegraphics[width=0.7\textwidth]{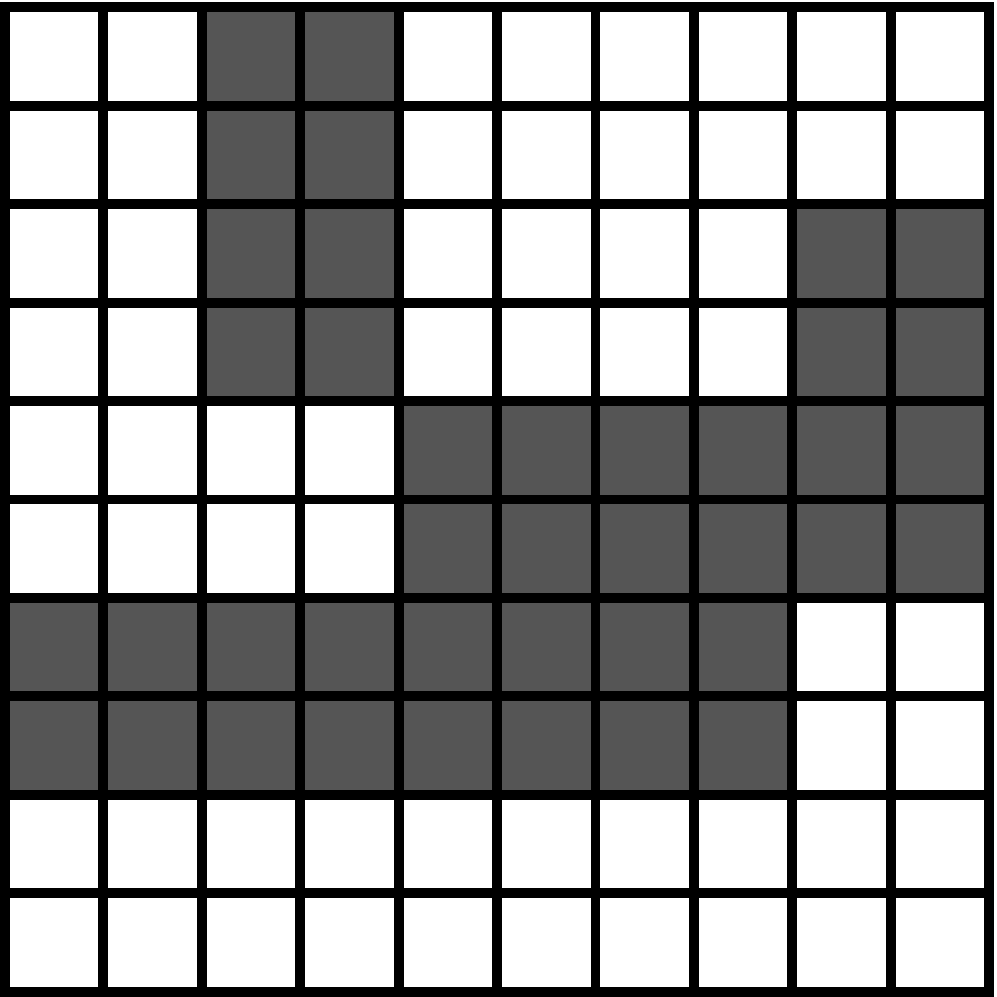}
    \caption{Resample Strategy}
  \end{subfigure}
  \begin{subfigure}[t]{0.32\columnwidth}
    \centering\includegraphics[width=0.7\textwidth]{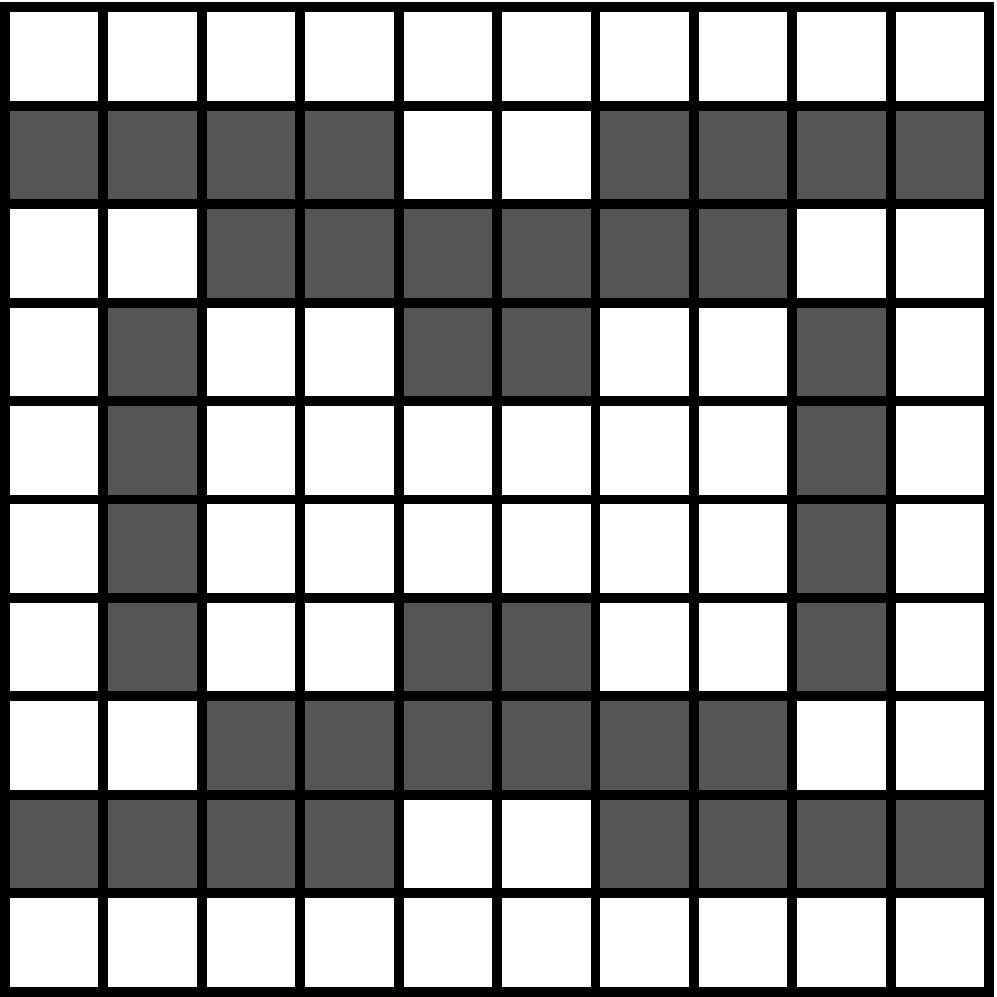}
    \caption{Mirror Strategy}
  \end{subfigure}
    \caption{
      An exemplary demonstration of the \textit{resample} (b) \textit{mirror} (c) strategies for increasing volume size on a (2D) dataset (a) to scale 2, i.e., doubling the volume size in each dimension.
      For both strategies, the foreground (grey pixels) still form a plausible foreground segmentation of a vessel network.
    }\label{fig:strategy_visualization}
\end{figure}

The demonstration of scalability requires some way of modifying a \textit{scale} parameter for a given dataset without changing other characteristics.
One possibility would be to use an already large dataset, scaling it down or clipping it to a smaller region.
However, besides the problem of availablility of a very large volume (representing the results of \textit{future} microscopes), this could cause artifacts and loss of detail due to downsampling, making the comparison of generated graphs very difficult.
Instead we use a \textit{small} (real world) volume and artificially increase its size using two strategies (see \autoref{fig:strategy_visualization}):

The \textit{resample} strategy: Resampling the volume with larger resolution, using nearest-neighbor sampling to avoid changing topology near very thin connections in the original volume.

The \textit{mirror} strategy: Repeating the volume in each dimension until the desired (integer) scale is achieved.
In order to generate a mostly connected network, in each dimension the $2i+1$'th volumes are mirrored.

In a real world scenario, improved acquisition technology would result in a combination of both increasing (voxel) size of previously visible vessels \textit{and} revealing previously undetected finer networks of larger complexity.
When not specified otherwise, the dataset \textit{Lymphatic 1} (\autoref{fig:datasets_lymphatic1}, $135\times160\times213$ voxels) was used for evaluation, as it has a non-tree-like topology, anisotropic voxel resolution (\SI{32x32x16}{\micro\metre}) and a irregular vessel shape.
In the following, \textit{scale} denotes the factor that each dimension of the dataset was multiplied with using one of the above strategies.
\begin{figure}[t]
    \includegraphics[width=\columnwidth]{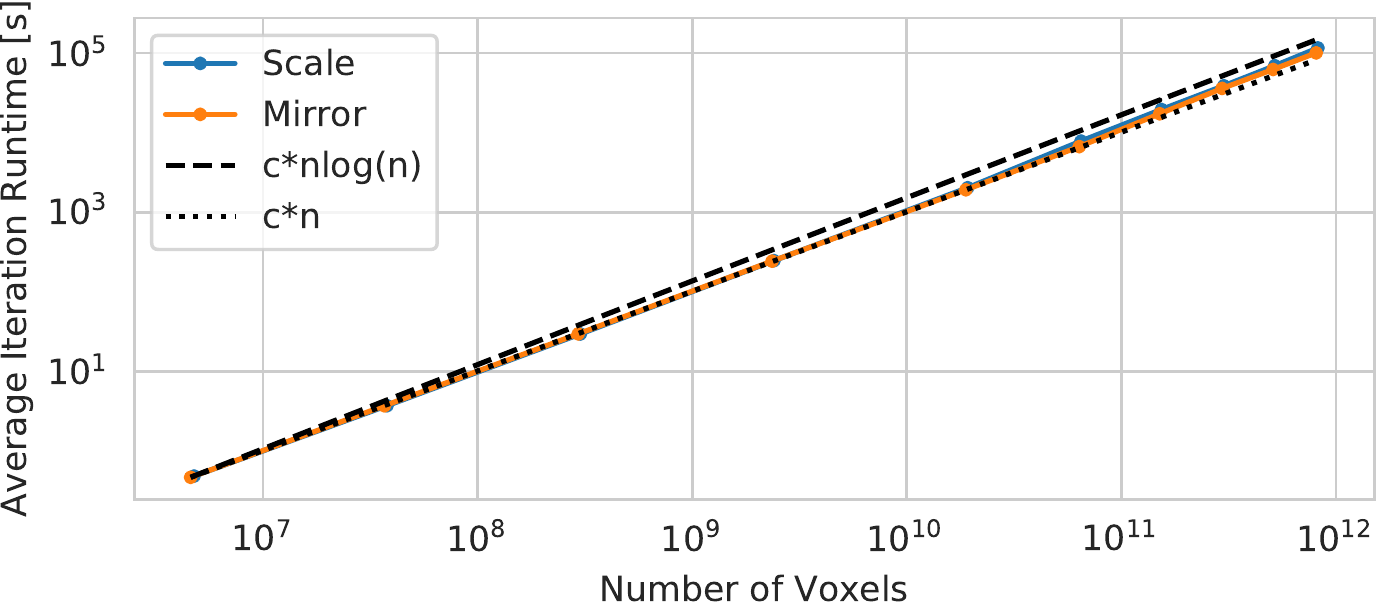}
    \caption{
      Average iteration runtime of the pipeline for the \textit{resample} and \textit{mirror} strategies in a log-log plot.
      The functions $c\cdot n\log n$ and $c\cdot n$ are shown as visual guides.
      $c$ was chosen so that both guides match the \textit{mirror} plot at the first data point.
    }\label{fig:runtime_loglog}
\end{figure}

\autoref{fig:runtime_loglog} shows the average runtime of a single refinement iteration of the pipeline depending on the number of voxels for the \textit{mirror} and \textit{resample} strategies in a log-log plot.
As shown, the runtime is only slightly worse than linear, as should be expected in the light of the derived runtime complexity of $\O(n \log n)$.
Furthermore, it should be noted that the increase in slope around $10^{11}$ coincides with the exhaustion of main memory and thus a larger number of disk accesses (cf.\ \autoref{fig:mem_rss}).

\begin{figure}[t]
  \begin{subfigure}[t]{\columnwidth}
    \includegraphics[width=\textwidth]{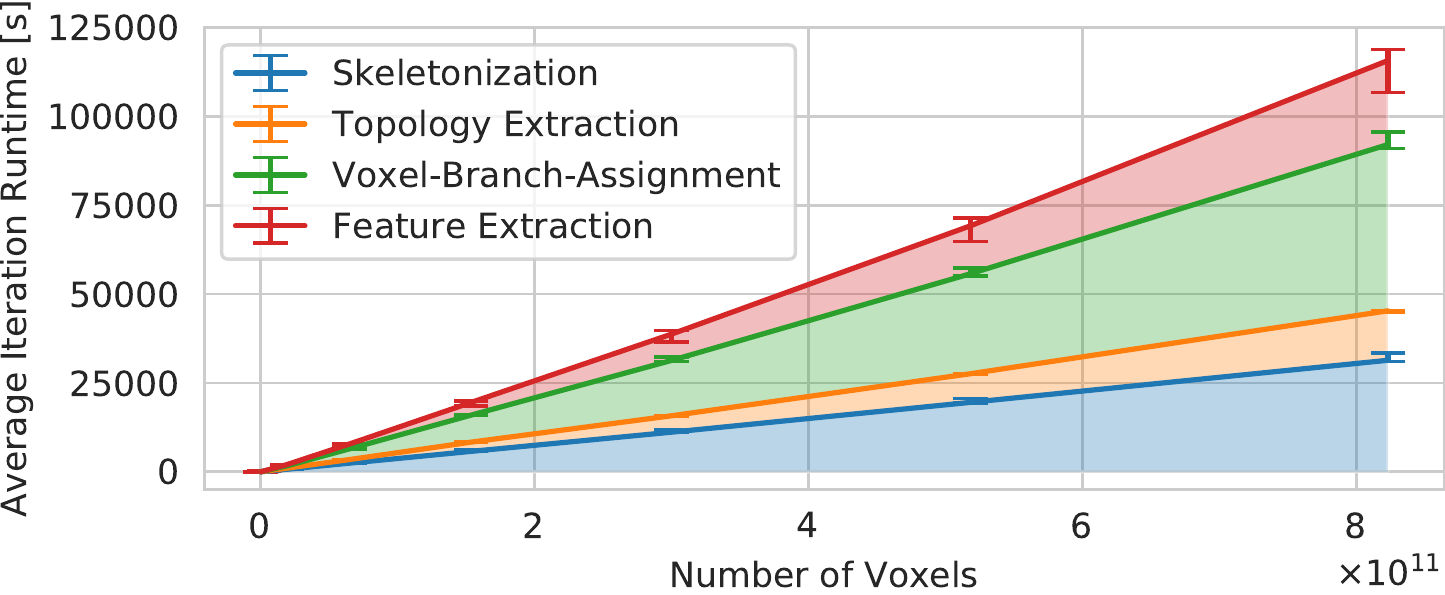}
    \caption{Resample Strategy}\label{fig:runtime_stack_scale}
  \end{subfigure}
  \begin{subfigure}[t]{\columnwidth}
    \includegraphics[width=\textwidth]{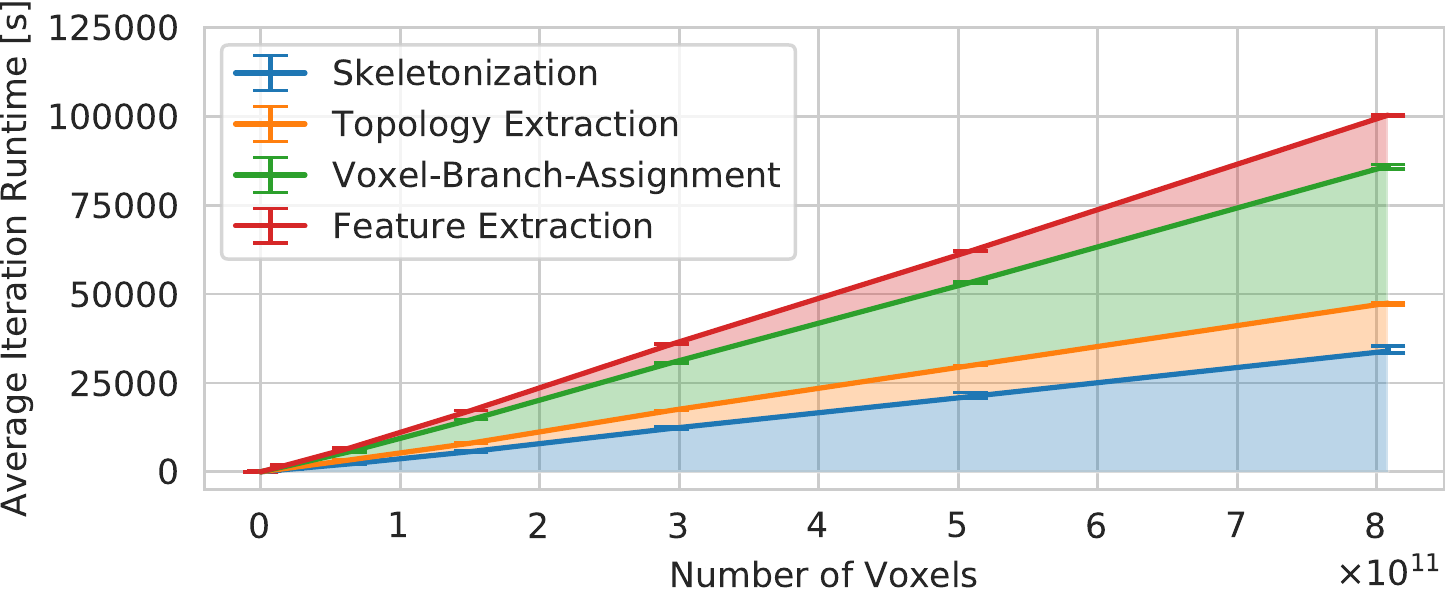}
    \caption{Mirror Strategy}\label{fig:runtime_stack_mirror}
  \end{subfigure}
    \caption{
      Iteration runtime of individual stages of the pipeline for the \textit{resample} (a) and \textit{mirror} (b) strategies.
      Note that the execution time of the Refinement step is not shown as it is very low compared to all other stages.
    }\label{fig:runtime_stack}
\end{figure}

\autoref{fig:runtime_stack} once again shows the average iteration runtime, but using a linear scale and with details of individual steps of the pipeline.
It can be observed that the runtime of all steps increases roughly linearly with respect to the number of voxels.
Here, it becomes apparent that for the \textit{resample} strategy, an iteration appears to take slightly more time than with the \textit{mirror} strategy.
This appears to be mostly due to more time spent in the Voxel-Branch-Assignment and Feature Extraction steps -- likely because larger vessels result in more time spent in searching the k-d-trees in both steps.

\begin{figure}[t]
    \includegraphics[width=\columnwidth]{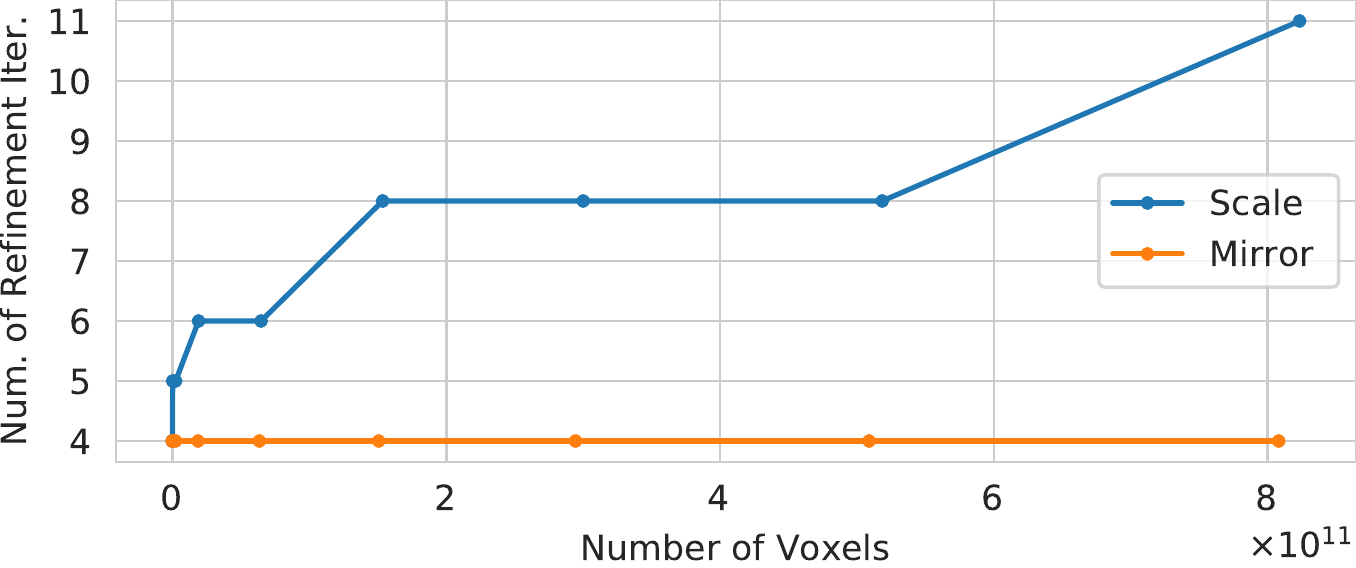}
    \caption{
      The total number of iterations required to reach a fixed point for the \textit{resample} and \textit{mirror} strategies.
    }\label{fig:runtime_numiter}
\end{figure}

The total runtime of the pipeline, however, is not only dependent on a single refinement iteration, but also on the number of iterations required to compute the final result.
The number of iterations required to reach a fixed point for an increasing number of voxels is depicted in \autoref{fig:runtime_numiter} for both scaling strategies.
While the total number of iterations using the \textit{resample} strategy can expected to be constant for scale~4 and higher (since scale~8 corresponds to 8 copies of the dataset at scale~4), the number of iterations remains at 4 independent of the scale in this experiment.
While we do not derive a maximum for the number of iterations for the \textit{resample} strategy, we can observe the required iterations to increase only moderately with the number of voxels in the experiment.
While we acknowledge that the (total) processing time of longer than one week for a 1 Terabyte dataset is certainly far from interactive use, this does not pose a problem in biomedical research since the time required for the preparation, image aquisition and postprocessing of a sample is of the same magnitude, and the presented pipeline is to the best of our knowledge the first to make analysis of these samples possible.
Furthermore, as our pipeline can be considered unbiased (see \autoref{sec:bulge_size}), interactive parameter tuning is not required.

\begin{figure}[t]
    \includegraphics[width=\columnwidth]{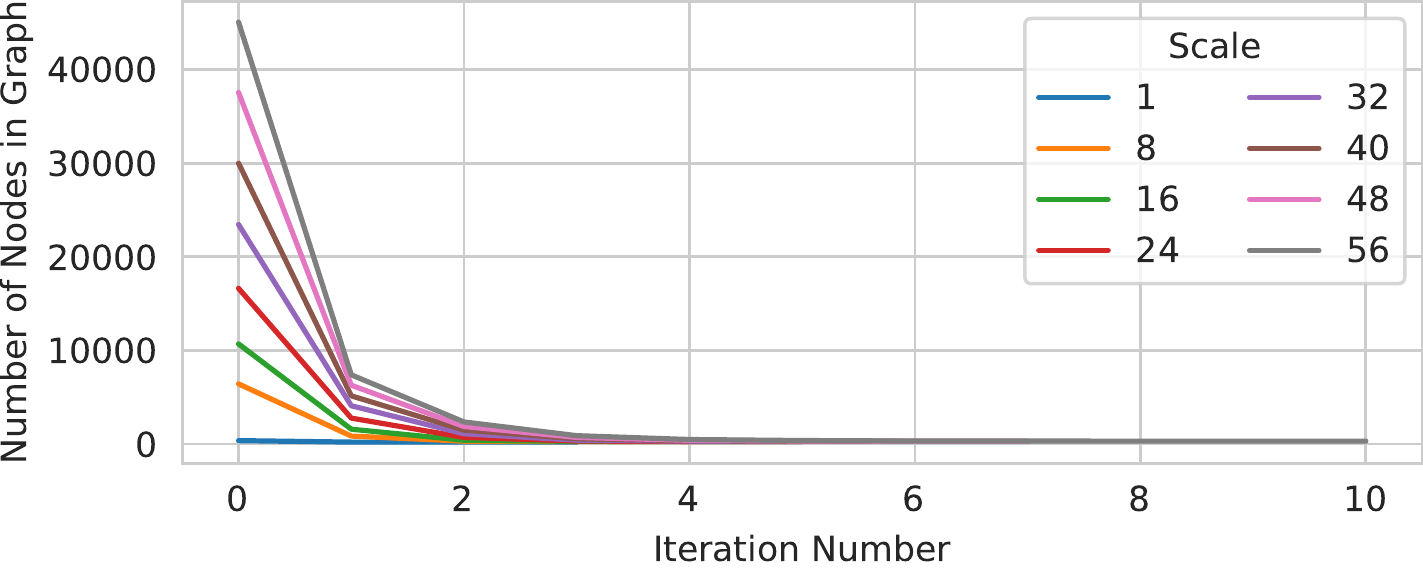}
    \caption{
      The total number of nodes in the graph after different numbers of iterations for the various scales.
    }\label{fig:numnodes_scale}
\end{figure}
\autoref{fig:numnodes_scale} shows that the total number of nodes in the graph rapidly decreases even for higher scales.
The final number of nodes for all scales is very close to scale~1, but increases slightly for higher iterations.
\autoref{sec:eval_image_resolution_invariance} and \autoref{fig:graph_comparison} demonstrate that the resulting graphs are actually similar.
Thus, if constant runtime (for different datasets of a given scale) is required, a relatively low maximum number of iterations can be specified to obtain a good approximation of running until a fixed point is reached.
At the same time, \autoref{fig:numnodes_scale} shows that a \textit{single} refinement step is not sufficient to obtain meaningful results for very large datasets, as a large number of spurious branches remain.
This underlines the necessity of the presented iterative refinement approach.

\subsection{Main Memory Scalability}
In the description of the pipeline we have argued for all steps of the pipeline to allocate $\O(m^{\frac{2}{3}})$ of memory.
For that reason and because total allocated heap memory is difficult to measure efficiently, we demonstrate how the implementation of the presented pipeline behaves with regard to the \textit{resident set size (RSS)}, which specifies the portion of the memory map of a process that is actually held in main memory.
This excludes allocated memory that has been swapped out to disk, but includes (sections of) files that have been copied to memory.
\begin{figure}[t]
    \includegraphics[width=\columnwidth]{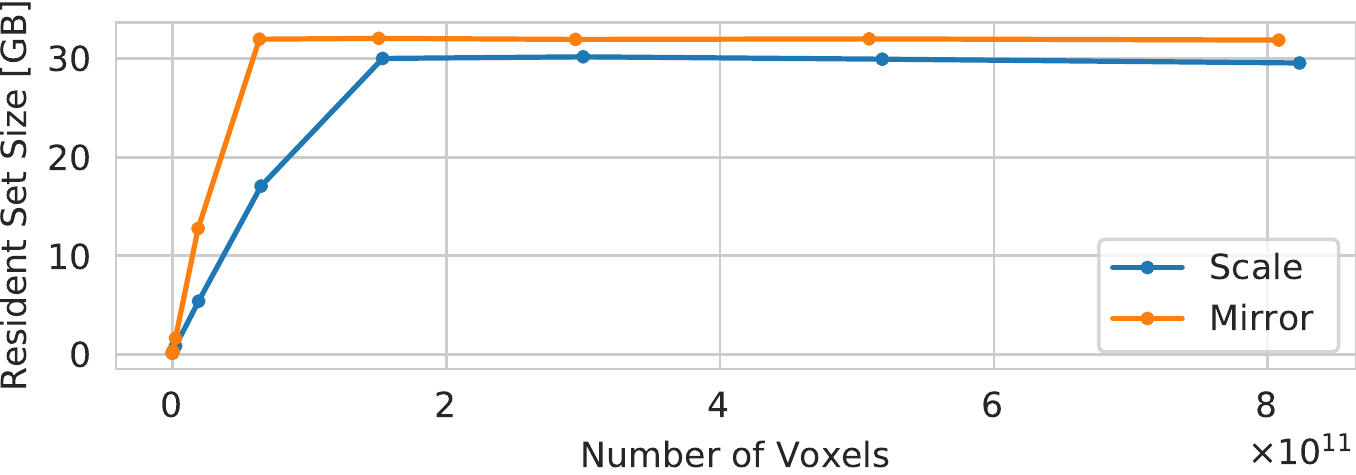}
    \caption{
      Maximum resident set size (RSS) of the graph extraction process for different problem sizes (number of voxels in the volume) for the \textit{resample} and \textit{mirror} strategies.
    }\label{fig:mem_rss}
\end{figure}
\autoref{fig:mem_rss} shows the maximum RSS of the process performing the graph extraction for different problem sizes (number of voxels in the volume).
It can be observed that the RSS increases for higher problem sizes, but does not exceed a limit near the total available main memory of the test machine (32GiB).
This demonstrates the aforementioned property of the pipeline to be scalable in terms of memory, but to use all of the available memory resources.
Notably, the \textit{mirror} strategy reaches the 32GiB limit earlier than the \textit{resample} strategy.
This may be due to the fact that by repeating the graph $m$-times in \textit{each} dimension, the total number of centerline points in the graph (and thus the required memory to store it) increases roughly by a factor of $m^3$, while for the \textit{resample} strategy, the number of centerline points is (roughly) multiplied by $m$.

\subsection{Image Resolution Invariance}\label{sec:eval_image_resolution_invariance}

\begin{figure*}[t]
  \begin{subfigure}[b]{0.32\textwidth}
    \centering\includegraphics[width=0.7\textwidth]{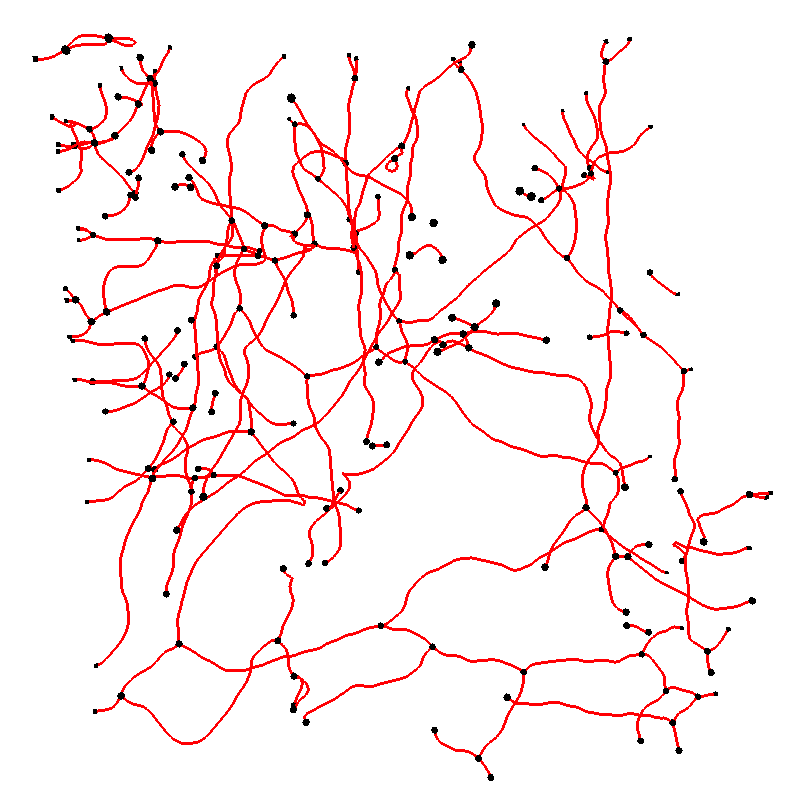}
    \caption{Scale 1, iterative refinement}\label{fig:graph_comparison_scale1_full}
  \end{subfigure}
  \begin{subfigure}[b]{0.32\textwidth}
    \centering\includegraphics[width=0.7\textwidth]{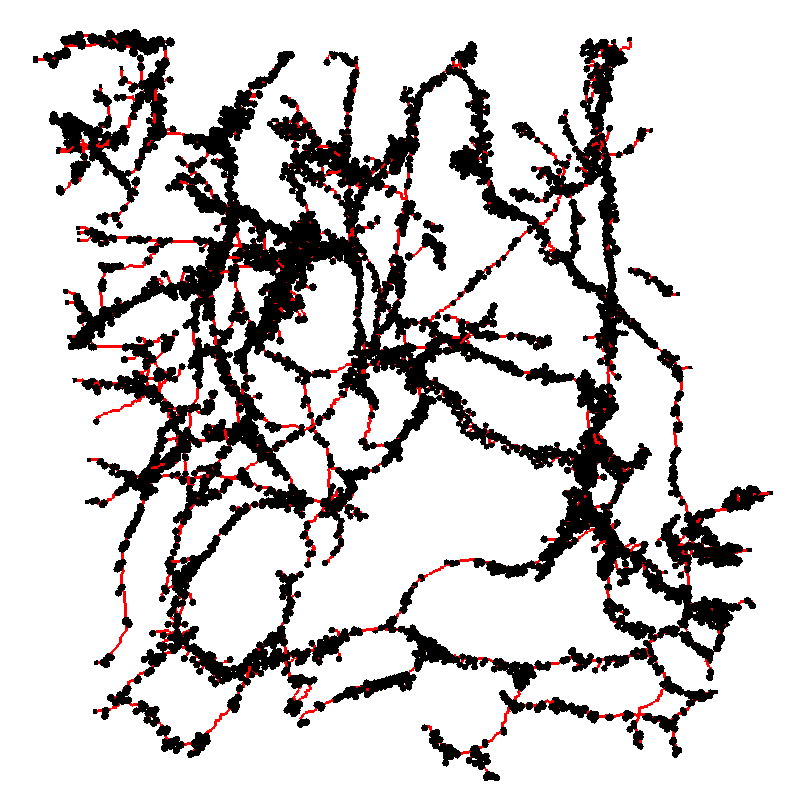}
    \caption{Scale 16, no refinement}\label{fig:graph_comparison_scale16_no}
  \end{subfigure}
  \begin{subfigure}[b]{0.32\textwidth}
    \centering\includegraphics[width=0.7\textwidth]{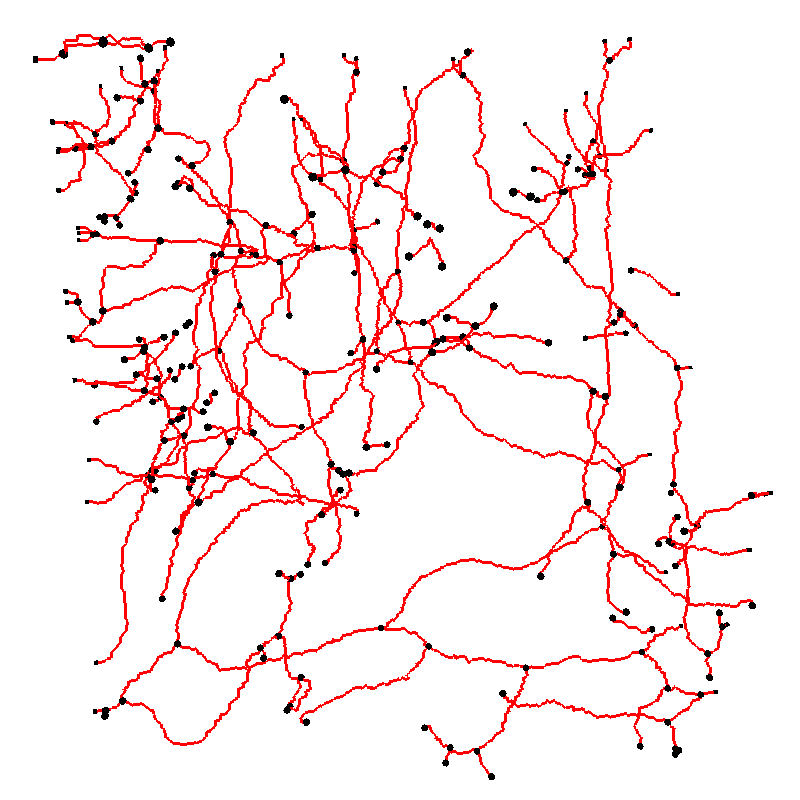}
    \caption{Scale 16, iterative refinement}\label{fig:graph_comparison_scale16_full}
  \end{subfigure}
    \caption{
      A demonstration of the effect of increased image resolution on the intermediate result and how the refinement iterations mitigate this effect, using dataset Lymphatic 1:
      An increased image resolution highly increases the number of erroneous branches (in this example the total number of branches is increased roughly 60-fold).
      After 5 refinement iterations the resulting graph is very similar to the graph extracted from the original volume without artificially increased resolution.
    }\label{fig:graph_comparison}
\end{figure*}
As a qualitative evaluation, \autoref{fig:graph_comparison} demonstrates the effect of increased image resolution on current skeletonization/graph extraction approaches and how the iterative refinement approach solves this problem.
For a 16-fold resolution increase in each of the coordinate axes (using the resample strategy), the number of branches and nodes increases sharply (\autoref{fig:graph_comparison_scale16_no}) compared to the the final graph extracted from the original volume (60-fold increase for the depicted example, \autoref{fig:graph_comparison_scale1_full}).
After 5 refinement iterations, the number of branches is reduced (\autoref{fig:graph_comparison_scale16_full}) to a number comparable to the simple case (scale 1, \autoref{fig:graph_comparison_scale1_full}).
As the increase in resolution allows for finer grained skeletonization, the graphs still differ in some details, but most of the structure is the same.
This is confirmed by biomedical domain experts who consider the refined version appropriate for further analysis, but reject the graph without refinement.

\begin{figure}[t]
    \includegraphics[width=\columnwidth]{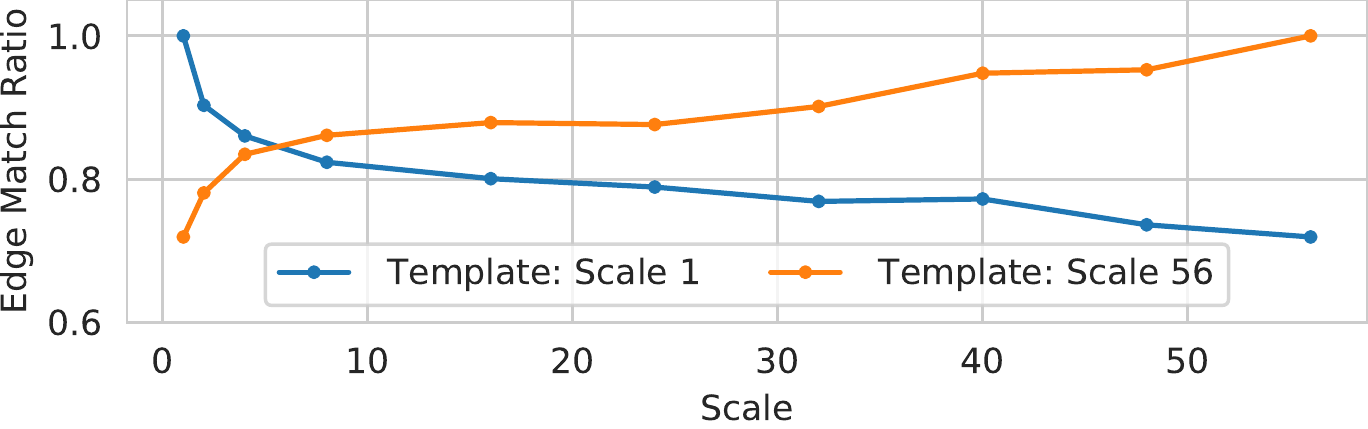}
    \caption{
      A similarity comparison between graphs of different scales using the \textit{edge match ratio} \cite{drees2019gerome} and the graphs extracted from the lowest and highest scale volume as template graph.
    } \label{fig:emr_scales}
\end{figure}
\autoref{fig:emr_scales} shows how an increase in resolution (using the \textit{resample} strategy) affects the extracted graph.
Since no ground truth graph is available for the dataset, we compare the all available graphs with the graph extracted using the \textit{lowest} and \textit{highest} scale, respectively.
When using the graph of scale 1 as a template, the edge match ratio drops relatively sharply for increasing, but smaller scales.
For higher scales the difference in similarity is not as pronounced.
An explanation could be that for low scales (and thus low image resolutions), skeletonization artifacts more heavily influence the final result.
Small bulges with a bulge size near the specified threshold cannot be treated with as much precision as necessary and are thus deleted although they should not be and vice versa.
This hypothesis is supported by the fact that when choosing the graph extracted from the \textit{highest} scale volume as the template for comparison, the graphs for most scales exhibit a much higher similarity.

\begin{figure}[t]
    \begin{subfigure}[t]{0.49\columnwidth}
      \centering\includegraphics[width=0.8\textwidth]{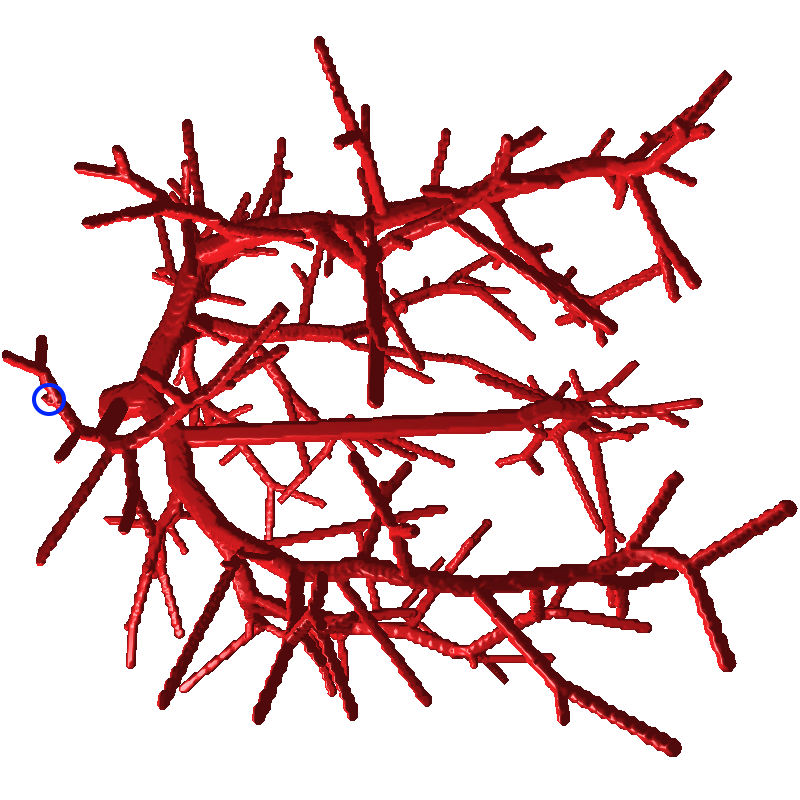}
      \caption{Synthetic 1 (no noise)}\label{fig:datasets_synthetic1}
    \end{subfigure}
    \begin{subfigure}[t]{0.49\columnwidth}
      \centering\includegraphics[width=0.8\textwidth]{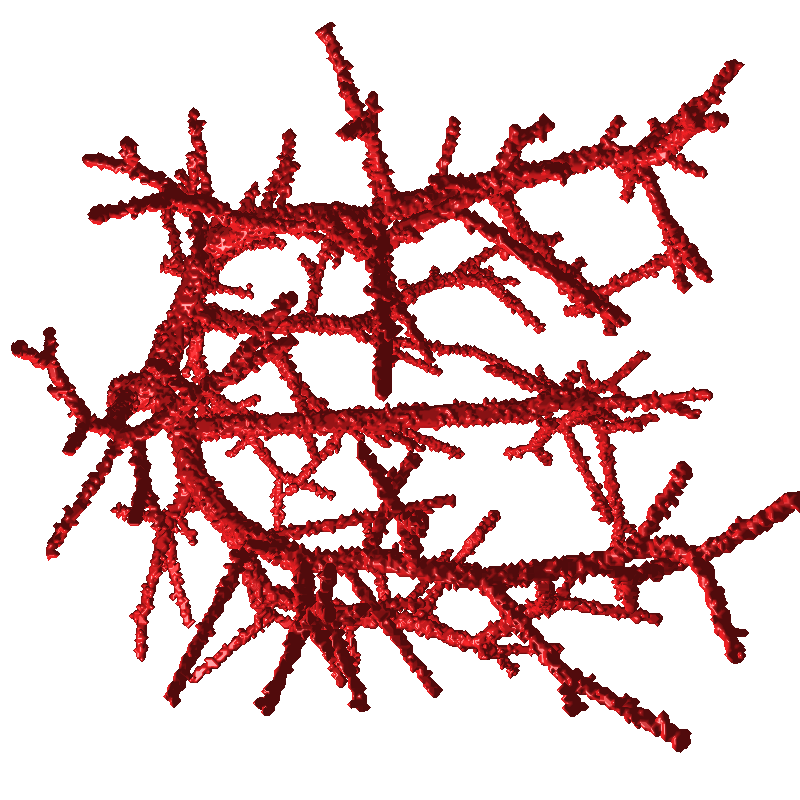}
      \caption{Synthetic 1 (noise level 0.2)}\label{fig:datasets_synthetic1_noise}
    \end{subfigure}
    \caption{
      The foreground of one of the synthetic vessel datasets used for evaluation without (a) and with maximal noise (b) applied.
      In (a) a very small bump which is categorized as a separate branch according to the ground truth is highlighted.
    }\label{fig:datasets_syn}
\end{figure}

\begin{figure}[t]
    \includegraphics[width=\columnwidth]{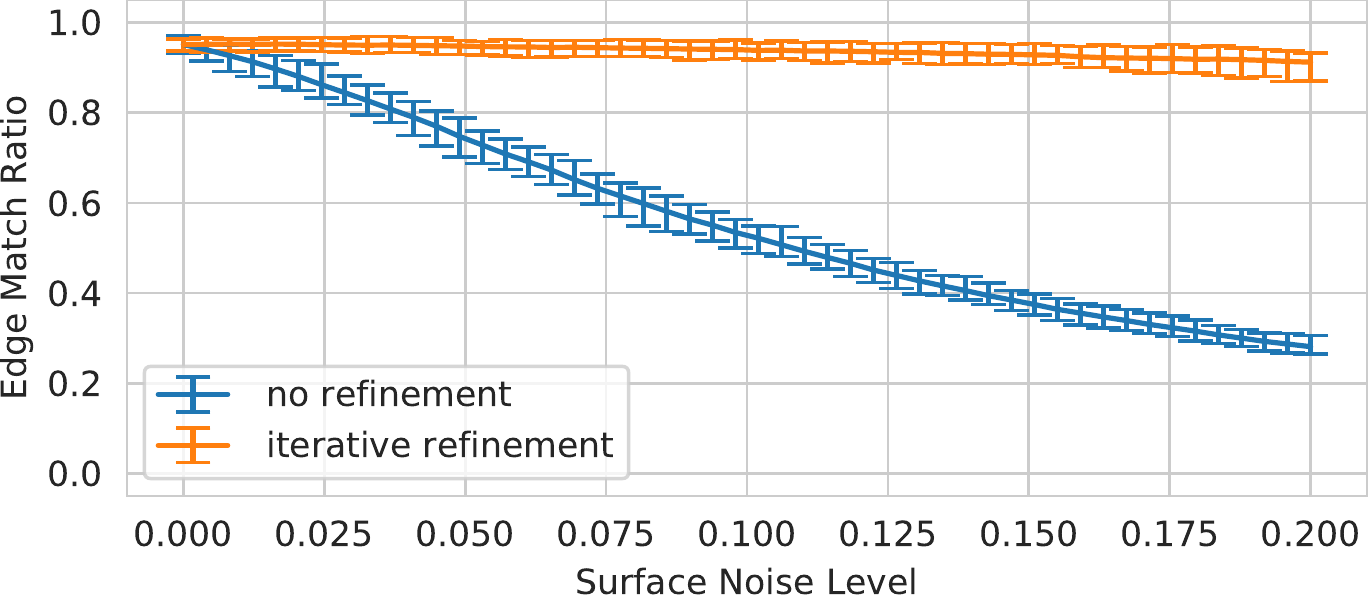}
    \caption{
      The edge match ratio~\cite{drees2019gerome} between the ground truth graph and the graph extracted from the corresponding volume~\cite{hamarneh2010vascusynth} with and without iterative refinement.
      Prior to the graph extraction salt-and-pepper noise is added to the surface (x-axis).
      For each noise level, minimum, maximum and average of 10 datasets for 4 surface noise seeds are shown.
    }\label{fig:noise_synthetic_match_ratio}
\end{figure}

As an additional way of evaluating the image resolution invariance (P3), we want to focus on one of the primary problem of increased image resolution:
The increase in surface noise on the individual voxel level.
For this we generate 10 datasets using Vascusynth~\cite{hamarneh2010vascusynth}, a tool for generating volumetric images of vascular trees as well as the corresponding ground truth segmentations and tree hierarchy used here.
Before graph extraction we perturb the surface of the binary volumes by iteratively selecting a random foreground or background surface voxel and flipping its value if this does not change the topology of the object.
The surface noise level is defined as $\frac{\text{\#voxel value changes}}{\text{\#total number of surface voxels}}$.
An example of a volume with applied surface noise is presented in \autoref{fig:datasets_synthetic1_noise}.
For each of the volumes, four variants with different random noise seeds were evaluated.
As shown in \autoref{fig:noise_synthetic_match_ratio}, the edge match ratio~\cite{drees2019gerome}, i.e., the ratio of edges that can successfully be matched between the extracted and the ground truth graph, rapidly decreases when not using iterative refinement while the refinement approach presented in this paper manages to reproduce the expected graph even for high noise levels with only slight decreases in similarity.
Note that we chose a bulge size of 1.5 for the iterative refinement, which is unusually low for blood vasculature.
However, Vascusynth generates some branches that are very short compared to the radius parent vessel, and in some cases are entirely enclosed, which are only visible as very shallow bumps (or not visible at all) in the generated volume (see highlighted region in \autoref{fig:datasets_lymphatic1} as an example).
For real world applications this is likely negligible.
Instead, higher robustness is likely preferable, necessitating higher bulge size, as discussed in the following section.

\subsection{Influence of the \texttt{bulge size} Parameter}\label{sec:bulge_size}
\begin{figure}[t]
  \begin{subfigure}[t]{0.49\columnwidth}
    \includegraphics[width=\textwidth]{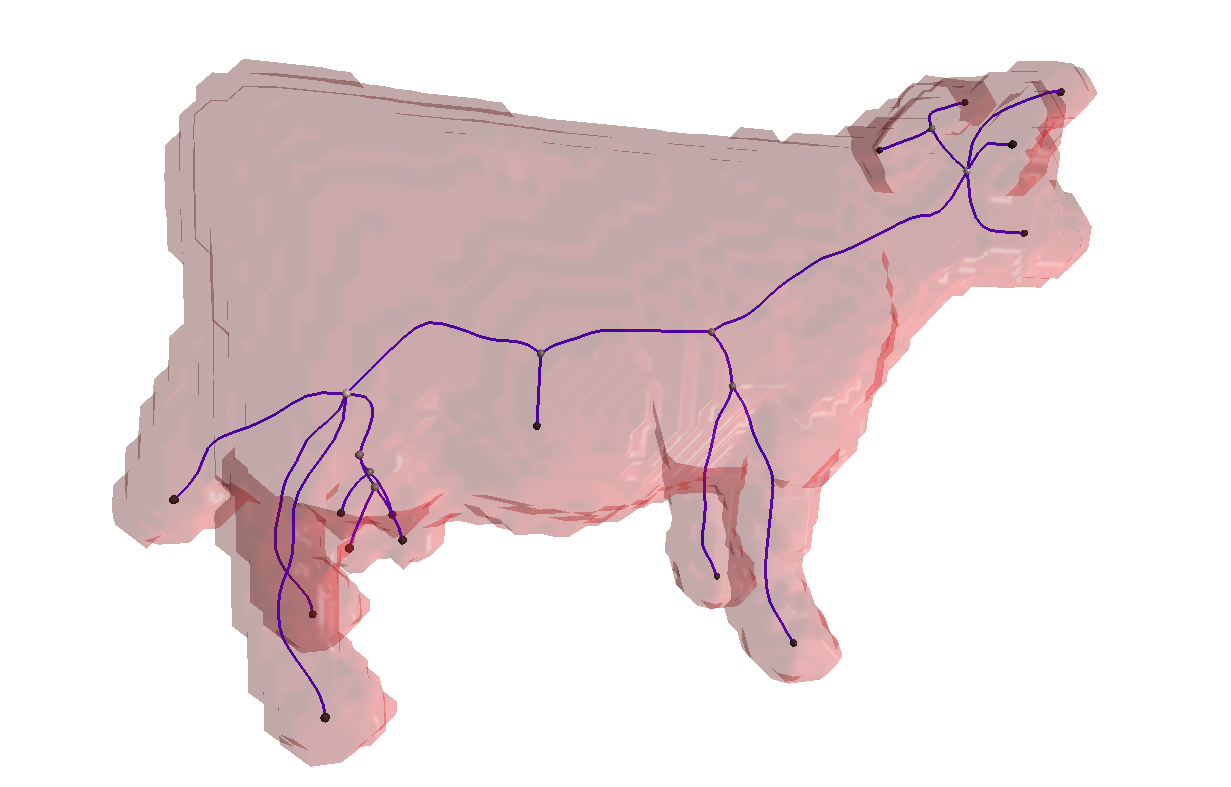}
    \caption{bulge size=0.2}
  \end{subfigure}
  \begin{subfigure}[t]{0.49\columnwidth}
    \includegraphics[width=\textwidth]{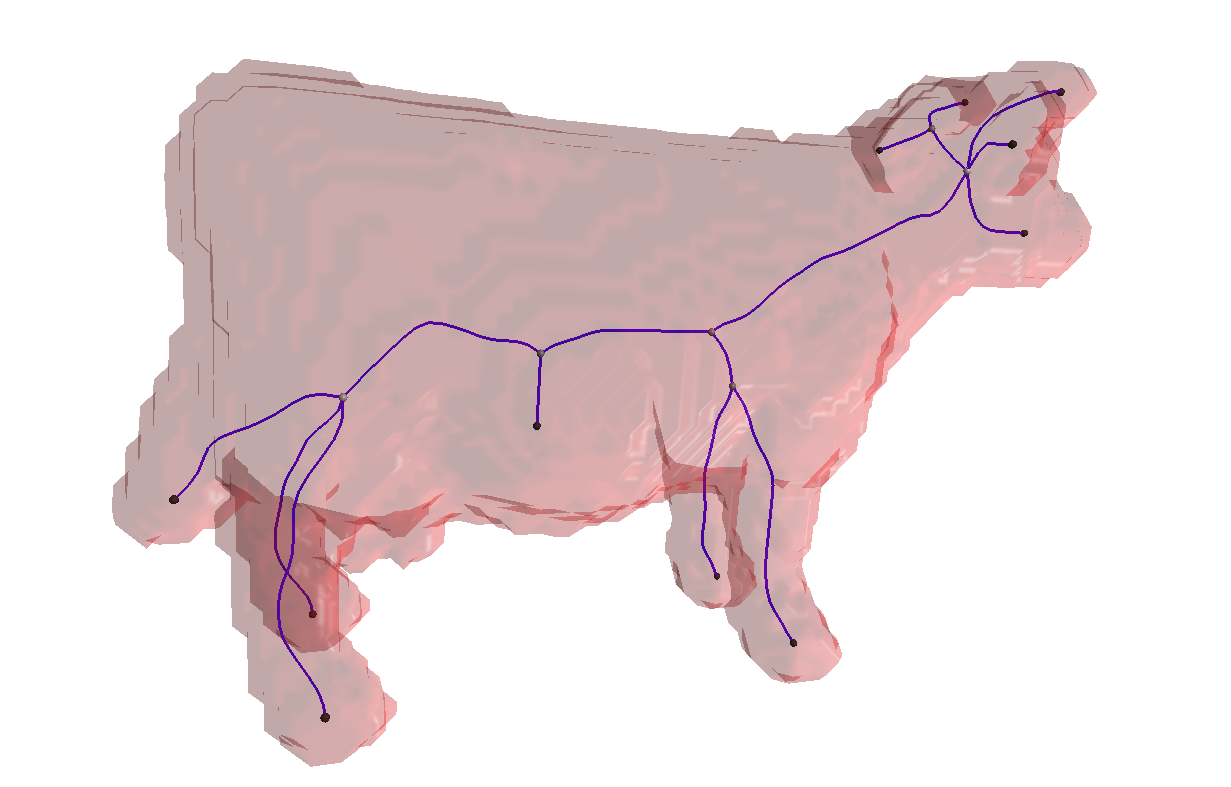}
    \caption{bulge size=0.3}
  \end{subfigure}
  \begin{subfigure}[t]{0.49\columnwidth}
    \includegraphics[width=\textwidth]{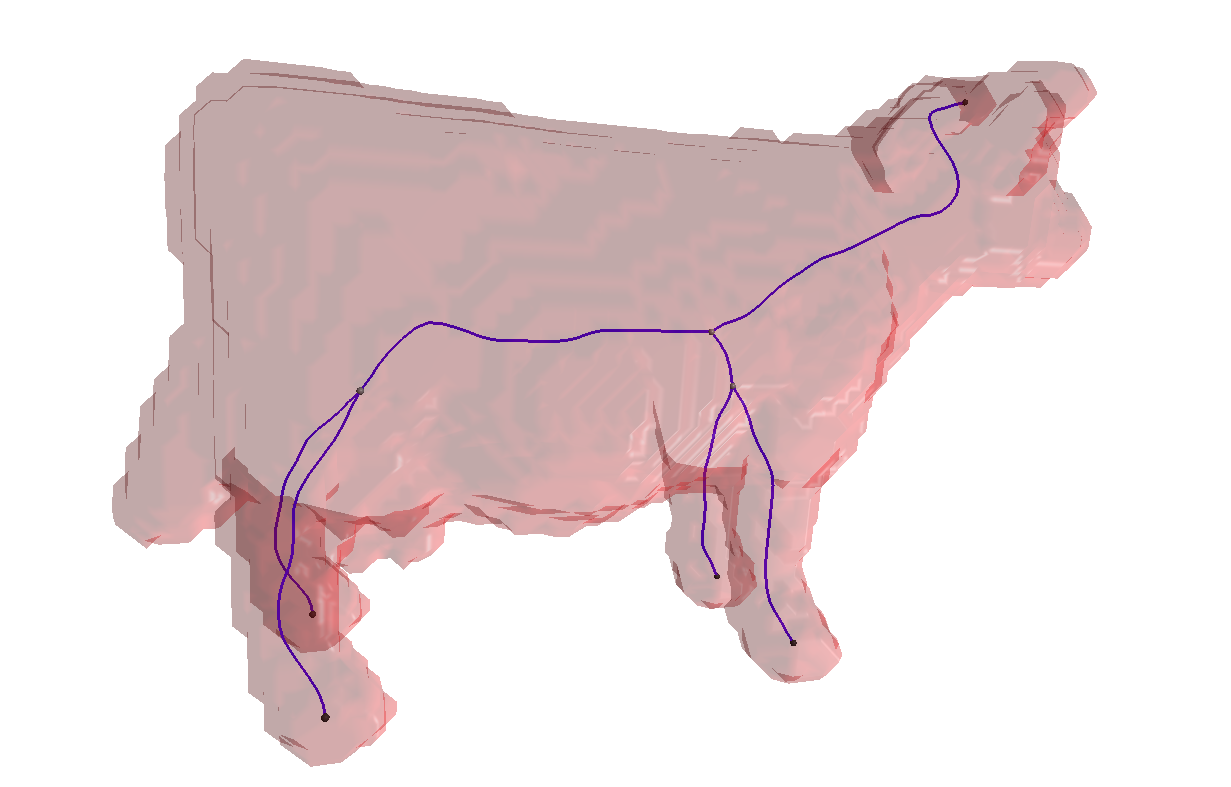}
    \caption{bulge size=1.0}
  \end{subfigure}
  \begin{subfigure}[t]{0.49\columnwidth}
    \includegraphics[width=\textwidth]{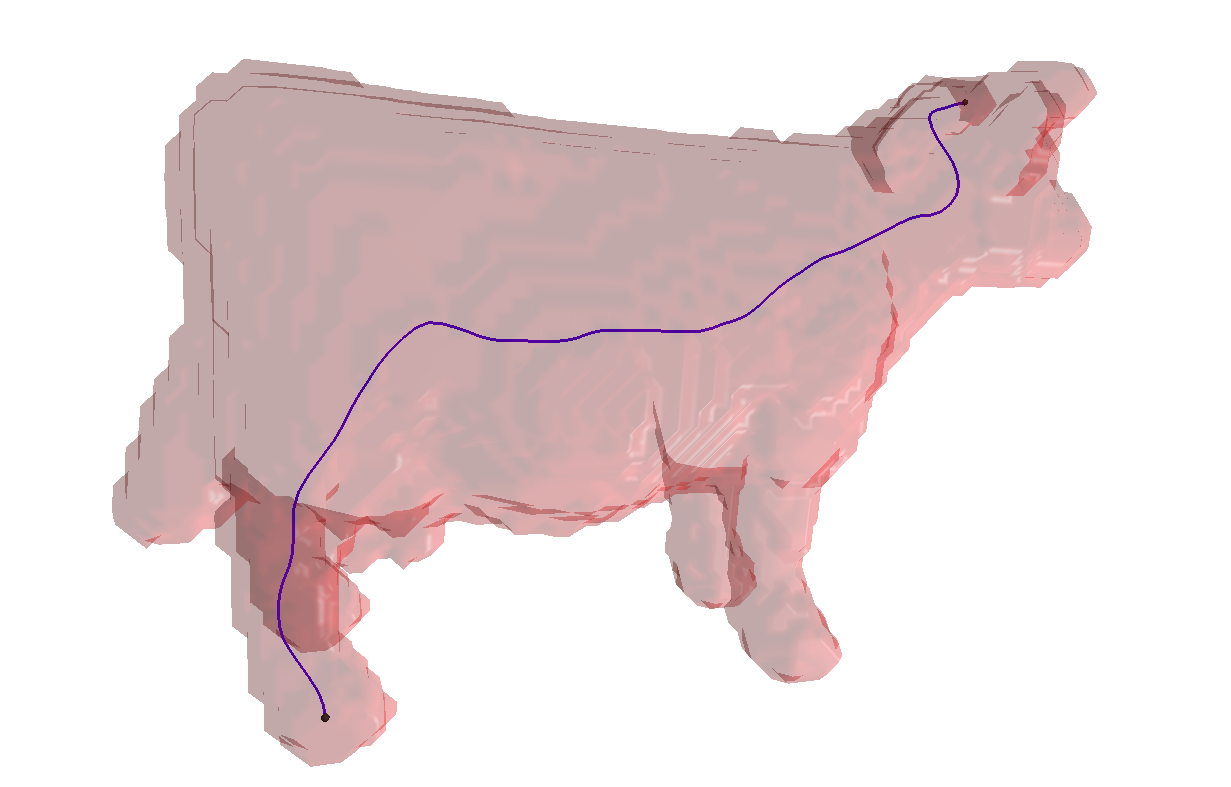}
    \caption{bulge size=3.2}
  \end{subfigure}
    \caption{
      A demonstration on how different parameter values affect the result of the graph extraction pipeline. In the example, very low values allow even very minor bumps (such as the cow's teats in the model) to be considered separate branches. Increasing values gradually remove further branches such as the belly, tail, and features of the head until finally for a very large value the graph consists of a single edge.
    }\label{fig:cows}
\end{figure}
\begin{figure}[t]
    \begin{subfigure}[t]{0.49\columnwidth}
      \centering\includegraphics[width=0.8\textwidth]{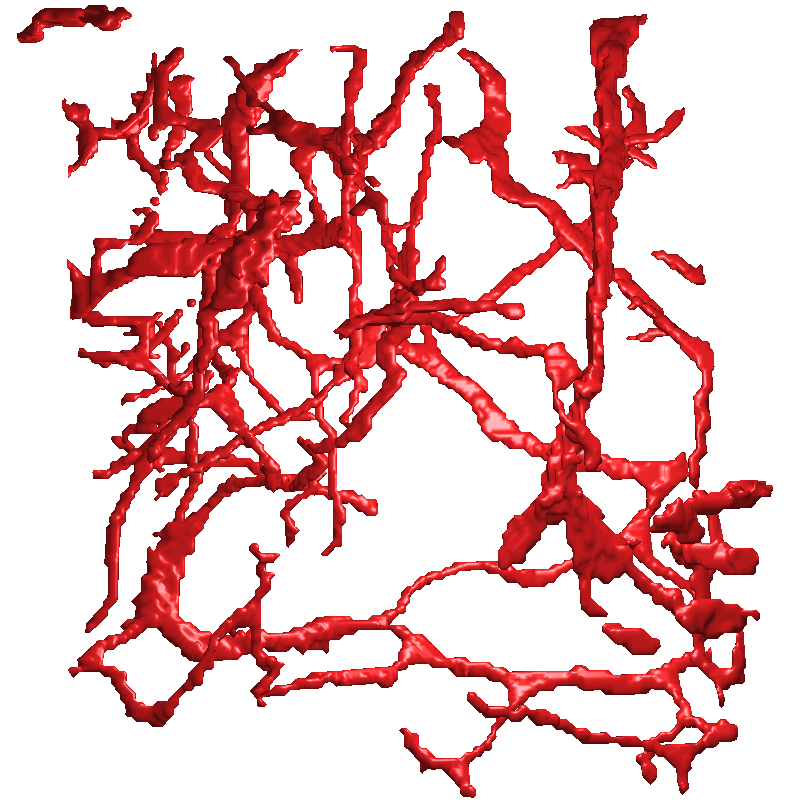}
      \caption{Lymphatic 1}\label{fig:datasets_lymphatic1}
    \end{subfigure}
    \begin{subfigure}[t]{0.49\columnwidth}
      \centering\includegraphics[width=0.8\textwidth]{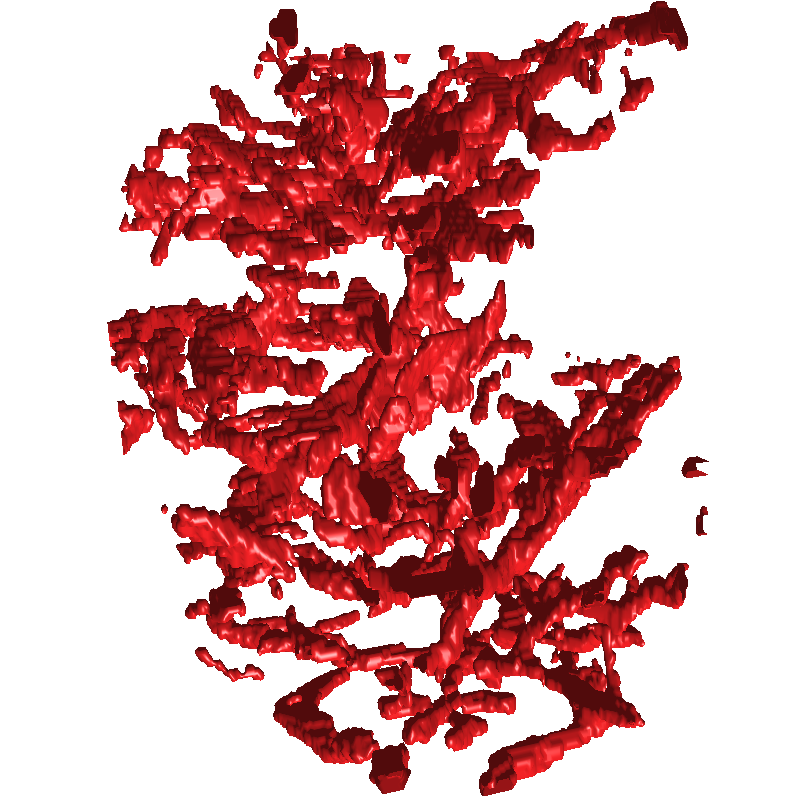}
      \caption{Lymphatic 2}\label{fig:datasets_lymphatic2}
    \end{subfigure}
    \caption{
      Some exemplary real world datasets used for evaluation rendered as the surface of the foreground.
      It is apparent that the surface and topology of the real world, lymphatic vessel datasets are much more complex than that of the synthetic blood vessel datasets (see \autoref{fig:datasets_syn}).
    }\label{fig:datasets_rw}
\end{figure}

As explained in \autoref{sec:refinement}, the parameter of the proposed method can be selected a-priori based on the application and the desired \textit{outcome} of the pipeline.
The \textit{quality} of the result does therefore not directly depend on the parameter.
Hence, we consider our method \textit{unbiased} (S1).
Nevertheless, in \autoref{fig:cows} we present an overview of the influence of different \texttt{bulge size} values on the result of the pipeline using a test dataset with low-profile, but variable-depth bulges.
As can be seen, a very low bulge size results in separate branches even for very small surface features.
Increasing the parameter gradually reduces the complexity of the extracted topology.
For real world applications, the bulge size should be set a-priori based on knowledge of the dataset.
For example, lymphatic vessels (especially in the pathological case, illustrated in \autoref{fig:datasets_lymphatic2}, but even in healthy individuals, see \autoref{fig:datasets_lymphatic1}) often have comparatively short branches, but also suffer from irregular surface features, often near branching points.
We therefore suggest a bulge size of 1.5, where the edge case corresponds to nearly spherical, but slightly elongated bulges.
On the other hand, (healthy) blood vasculature often exhibits a smooth surface, round diameter, and clearly defined branches, so that a bulge size of 3.0 or higher can be chosen confidently.

\subsection{Anisotropic Resolution}
\begin{figure}[t]
  \includegraphics[width=\columnwidth]{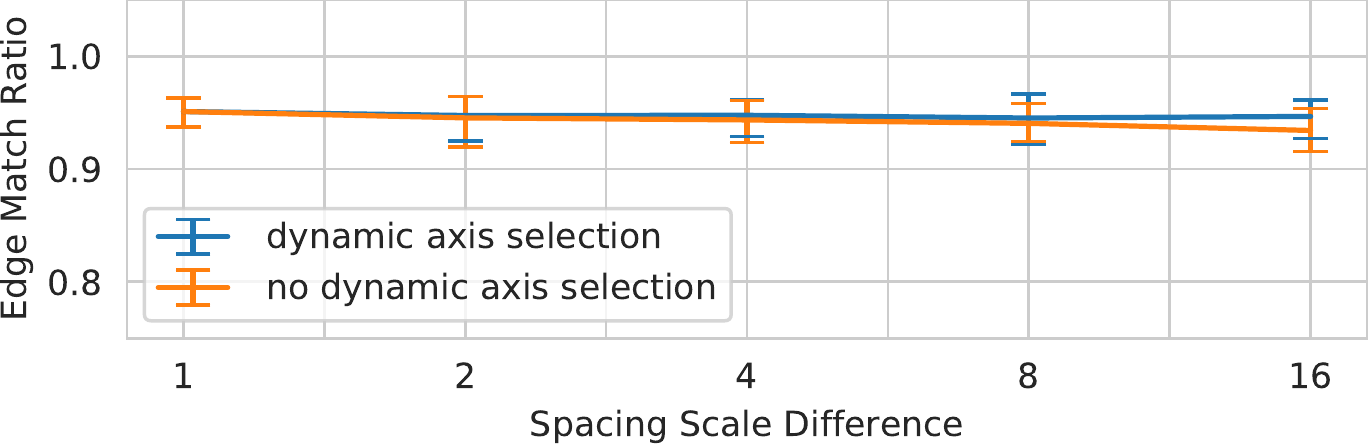}
  \includegraphics[width=\columnwidth]{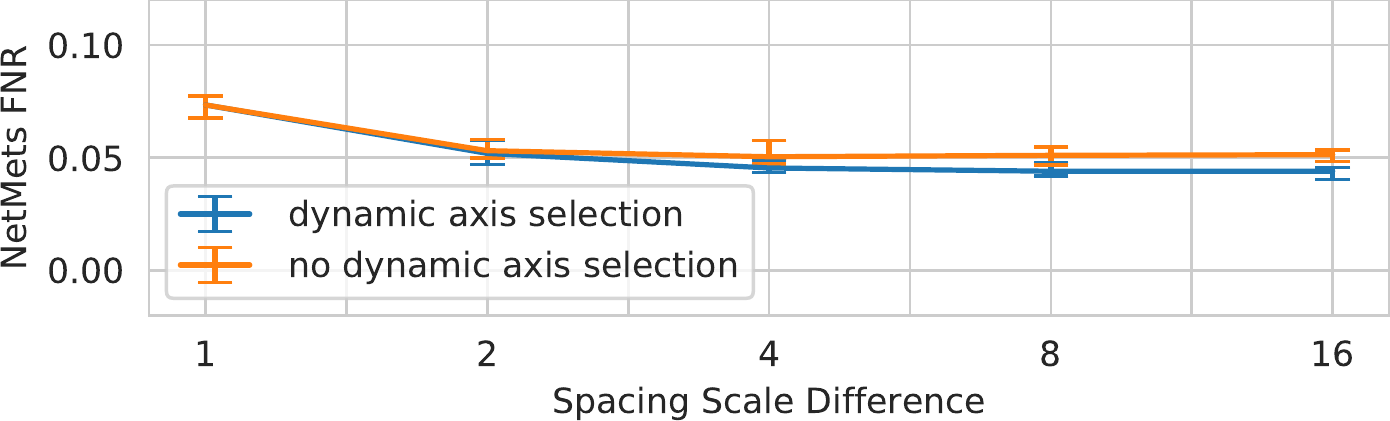}
  \includegraphics[width=\columnwidth]{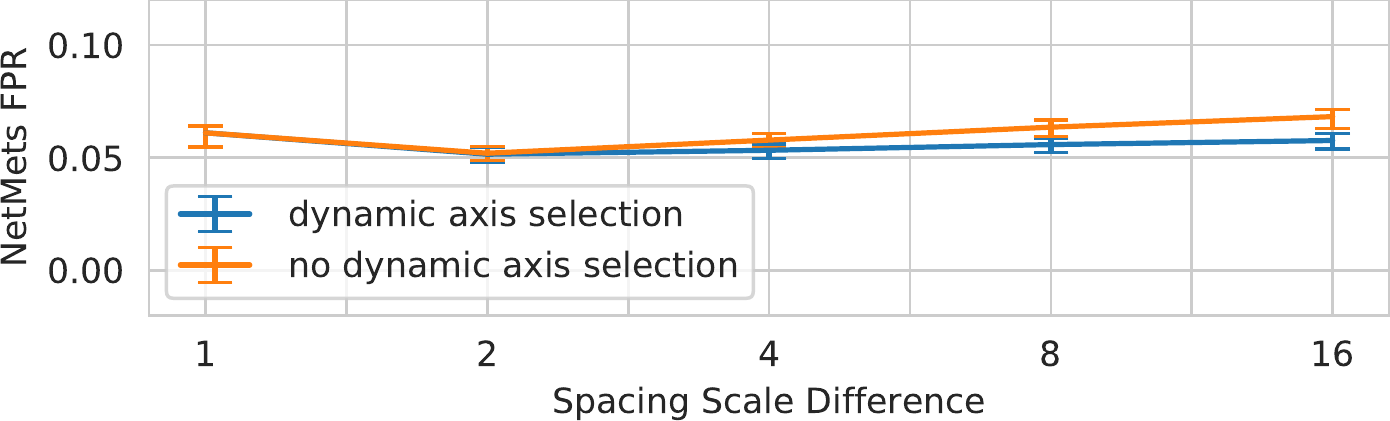}
  \caption{
    A demonstration of how anisotropy affects the topological (Edge Match Ratio~\cite{drees2019gerome}) and geometrical (Centerline~\cite{mayerich2012netmets}) structure of the extracted graph. We compare the ground truth graph of a volume generated by VascuSynth~\cite{hamarneh2010vascusynth} to the graph extracted using the proposed pipeline after increasing the x- and y-resolution by a specified factor (but leaving z-axis resolution unchanged) and resampling the volume. Minimum, average and maximum for 10 datasets are shown. As can be seen, anisotropic resolution does not strongly affect the pipeline and the dynamic axis selection is advantageous.
  }\label{fig:anisotropy_synthetic}
\end{figure}

The third secondary requirement states that a pipeline should be able to operate on volumes with anisotropic resolution (S3).
In order to evaluate our proposed pipeline in terms of that property, we use a volume with known ground truth and isotropic voxel spacing.
We then artificially create datasets with different, anisotropic resolutions by resampling the dataset in specific dimensions.
As an example, confocal microscopes due to their anisotropic point spread function generate volumes with a comparatively low z-resolution, which is why in this experiment we chose to leave the z-axis resolution intact and increase the resolution in x- and y-direction (akin to the \textit{resample} strategy).
In total, we evaluated 10 datasets with resolution scale differences of 1, 2, 4, 8, and 16.
We compared the extracted topology using the graph matching method and edge match ratio similarity measure proposed in~\cite{drees2019gerome} and the quality of the extracted centerline geometry using the NetMets framework~\cite{mayerich2012netmets}.
In the case of NetMets the average radius of all edges multiplied by 10 was chosen as the parameter~$\sigma$.
The results are shown in \autoref{fig:anisotropy_synthetic}.
Although the ground truth is not matched perfectly and some drift between scales is noticeable (see discussion about ground truth quality above), in general no trend towards positive or negative impact on the evaluated score can be observed based on the level of anisotropy.
Furthermore, it should be noted that the dynamic axis selection discussed in \autoref{sec:skeletonization} indeed improves the result for larger anisotropy, although for smaller levels there is next to no difference between both variants.
We therefore conclude that the proposed pipeline is in fact able to process data of even highly anisotropic resolution.

\subsection{Robustness against shape deviations}

For the evaluation of robustness of the pipeline, especially with regard to deviations from the tube (S2) shape of analyzed vasculature, we use the robustness test introduced in~\cite{drees2019gerome}.
Additionally, the FNR and FPR values (false negative/positive ratio) of~\cite{mayerich2012netmets} were used to measure the geometric error using in the framework of~\cite{drees2019gerome}.
The (real world) lymphatic and synthetic blood vessel datasets were rotated and resampled (with doubled resolution in each axis) in \ang{10} steps around the $z$-axis, processed using the proposed pipeline, compared to the template graph.
The minimum similarity of all steps denotes the robustness index.
For the synthetic datasets the ground truth graph provided by the tool was used as a template and thus are also measuring the accuracy of the method.
For real world datasets, generation of an annotated ground truth graph is virtually impossible~\cite{drees2019gerome} and thus only the robustness can be evaluated.
We compare the results without refinement to the graphs extracted using a bulge size of 1.5 and iteration until reaching a fixed point.

\begin{table*}
  \begin{center}
    \begin{tabular}{lllllll}
\toprule
            & Property & $\mathcal{G}_{length}$ & $\mathcal{G}_{roundnessMean}$ & $\mathcal{G}_{straightness}$ & $\mathcal{N}_{FNR}$ & $\mathcal{N}_{FPR}$ \\
Dataset & Refinement &                        &                               &                              &                     &                     \\
\midrule
\multirow{2}{*}{Lymphatic 1} & Iterative Refinement &         \textbf{0.781} &                \textbf{0.780} &               \textbf{0.823} &     \textbf{0.0345} &     \textbf{0.0355} \\
            & No Refinement &                  0.450 &                         0.500 &                        0.578 &              0.0515 &              0.0505 \\
\cline{1-7}
\multirow{2}{*}{Lymphatic 2} & Iterative Refinement &         \textbf{0.736} &                \textbf{0.775} &               \textbf{0.785} &     \textbf{0.0328} &     \textbf{0.0341} \\
            & No Refinement &                  0.486 &                         0.536 &                        0.614 &              0.0550 &              0.0539 \\
\cline{1-7}
\multirow{2}{*}{Lymphatic 3} & Iterative Refinement &         \textbf{0.725} &                \textbf{0.760} &               \textbf{0.762} &     \textbf{0.0431} &     \textbf{0.0386} \\
            & No Refinement &                  0.487 &                         0.516 &                        0.612 &              0.0653 &              0.0632 \\
\cline{1-7}
\multirow{2}{*}{Synthetic 1} & Iterative Refinement &         \textbf{0.862} &                \textbf{0.600} &               \textbf{0.916} &              0.0445 &              0.0477 \\
            & No Refinement &                  0.623 &                         0.446 &                        0.696 &              0.0408 &              0.0488 \\
\cline{1-7}
\multirow{2}{*}{Synthetic 2} & Iterative Refinement &         \textbf{0.819} &                \textbf{0.598} &               \textbf{0.895} &              0.0425 &              0.0461 \\
            & No Refinement &                  0.603 &                         0.452 &                        0.690 &              0.0393 &              0.0479 \\
\bottomrule
\end{tabular}

  \end{center}
  \caption{
    Results of the robustness test~\cite{drees2019gerome} using 36 rotations around the z-axis for each respective dataset.
    $\mathcal{G}_{property}$ denotes the GERoMe index for the given \textit{property}.
    $\mathcal{N}_{FNR}$ and $\mathcal{N}_{FPR}$ use the same perturbation procedure, but measure the geometric error~\cite{mayerich2012netmets} of the calculated centerlines and therefore show the aggregated maximum value.
    Refined graphs were extracted using a bulge size of 1.5 and iteration until reaching a fixed point.
    Relative score differences above 10\% are highlighted as bold text.
  }\label{tab:robustness}
\end{table*}

The aggregated results in \autoref{tab:robustness} show that for all properties the refinement procedure results in significantly higher robustness for all datasets.
As should be expected, the total robustness of the much more complex real world lymphatic vessel datasets is overall lower than for the (simpler) synthetic datasets.
However, it is noteworthy that even when applied to complex real world datasets, the proposed pipeline achieves a higher score than the simple, but commonly applied approach without refinement on simple, synthetic datasets.

\subsection{Remaining Secondary Requirements}
The fourth secondary requirement (the ability to process volumes of arbitrary topology) directly follows from how the topology is extracted from the binary skeletonized volume (\autoref{sec:topology_extraction}) and, in particular, is not restricted to a tree topology, in contrast to other methods~\cite{drechsler2010hierarchical}.
Furthermore, the pipeline is independent of the imaging conditions (S5) by operating on an existing foreground segmentation.

\subsection{Limitations}\label{sec:limitations}
The proposed pipeline -- like all topological thinning-based methods -- by definition suffers from distortions in the binary input image that causes changes to the topology, i.e., cavities in foreground objects and loops on the boundary.
While these features are not expected in the actual vessel structure, imaging artifacts, noise or problems with the segmentation procedure can still produce such artifacts in practice.
However, careful post-processing of the segmentation can mitigate these effects:
Cavities in foreground objects can reliably be removed using a modified variant of~\cite{isenburg2009streaming} operating on the \textit{background} without labeling, but removing objects smaller than a specified size.
The threshold can typically be chosen very liberally, e.g., as a percentage of the volume diagonal.
Removal of loops on the surface is more difficult, but in our experience a (binary) median filter performs well.
The filter size should be chosen to not disturb the smallest vessels in the image, but still be able to cover surface loops in the image.

\section{Conclusion}
Analyzing ultra large datasets by scalable algorithms is still a huge challenge. 
We have presented a pipeline for extracting the topology and various features of vessel networks from large three-dimensional images.
We were able to show that our method fulfills all requirements identified in \autoref{sec:requirements}.
Our main contribution, a novel iterative refinement approach and careful engineering of all pipeline stages, allowed us to demonstrate the scalability and thus applicability to very large datasets, e.g., generated by modern microscopes (primary requirements).
At the same time, our pipeline can be applied to a wide range of problem domains due to its robustness, unbiased nature, and lack of assumptions about the topology and morphology of the analyzed vessel systems (secondary requirements).

We will now apply the proposed pipeline and continue previous work~\cite{rene2017} using previously unachieveable level of detail, which hopefully leads to new biomedical discoveries.

In the future, we would like to explore how even more specific image features (c.f.~\cite{rodriguez2006rayburst}) can be integrated into the pipeline without compromising its scalability.
Additionally, the current version of the pipeline is entirely single-threaded.
While it is possible to more efficiently use the available resources of modern hardware by simultaneously processing multiple datasets in separate processes, it would be desirable to accelerate a single run using multiple processor cores or even GPUs.
However, this is a challenging task:
While approaches to parallel skeletonization algorithms exist, they pose problems with regard to guarantees about the mediality of the centerline and reproducibility in general.
Furthermore, at least in the current formulation, the connected-component-analysis~\cite{isenburg2009streaming}, which is used in variations in several places in this paper is inherently sequential.
Offloading work to the GPU requires even more attention to detail with regard to memory management.
Furthermore, we would like to explore automatic segmentation approaches for vessel structures in large volumetric datasets, with special focus on (irregular) lymphatic vessel systems, to facilitate the usability of the presented pipeline even further.

The presented pipeline is part of version 5.1 of Voreen~\cite{meyer2009voreen} a widely-used (\cite{landell2019material,benz2019low}) open-source volume processing and rendering framework.
All scripts used in the evaluation are publicly available online\footnote{\url{https://zivgitlab.uni-muenster.de/d_dree02/graph_extraction_evaluation}} to facilitate reproducing the presented results.

\section*{Acknowledgment}
This work was funded by the Deutsche Forschungsgemeinschaft (DFG) – CRC 1450 – 431460824.

\ifCLASSOPTIONcaptionsoff
  \newpage
\fi



%

\bibliographystyle{IEEEtran}
\bibliography{src}

%





\end{document}